\documentclass[preprint,review,10pt]{elsarticle}
\usepackage[english]{babel}
\usepackage[T1]{fontenc}
\usepackage{times}
\usepackage{etoolbox}
\makeatletter

\patchcmd{\ps@pprintTitle}{\footnotesize\itshape
      Preprint submitted to \ifx\@journal\@empty Elsevier
      \else\@journal\fi\hfill\today}{\scriptsize{Preprint submitted to ASOC \hfill \today}}{}{}

\usepackage{float}
\usepackage{natbib}
\usepackage{graphicx}
\usepackage[lined,  boxed,  linesnumbered,  ruled]{algorithm2e}
\graphicspath{{Figures/}}
\usepackage{lineno,hyperref}
\usepackage{amssymb, amsfonts, amsmath, graphicx, epstopdf, bbding, lipsum, multirow, makecell, picinpar, enumerate}
\usepackage[table]{xcolor}
\usepackage{colortbl}
\usepackage{xspace}
\usepackage{subfigure} 
\usepackage{xcolor}
\usepackage[normalem]{ulem} 

\makeatother
\usepackage{comment}
\usepackage{hyperref}
\usepackage{amsmath}

\usepackage{amsmath,amssymb,amsfonts}
\usepackage{algorithmic}
\usepackage[]{graphicx}
\graphicspath{{Figures/}}
\usepackage{lipsum}
\usepackage{balance}
\usepackage{textcomp}
\usepackage{xcolor}
\usepackage{multirow}
\usepackage{color}
\usepackage{lscape}
\usepackage{longtable,booktabs}

\begin{document}

\begin{frontmatter}

\title{RAINER: A Robust Ensemble Learning Grid Search-Tuned Framework \\ for Rainfall Patterns Prediction}

\author[add1]{Zhenqi~Li\corref{contrib}}
\ead{zl3508@columbia.edu}
\author[add3]{Junhao~Zhong\corref{contrib}}
\ead{jz455@duke.edu}
\author[add7]{Hewei~Wang}
\ead{hewei.wang@ucdconnect.ie}
\author[add10]{Jinfeng~Xu}
\ead{jinfeng@connect.hku.hku}
\author[add7]{Yijie~Li}
\ead{yijie.li@ucdconnect.ie} 
\author[add2]{Jinjiang~You}
\ead{jinjiany@andrew.cmu.edu} 
\author[add11]{Jiayi~Zhang}
\ead{smyjz19@nottingham.edu.cn}
\author[add4]{Runzhi~Wu}
\ead{wurunzhi@tju.edu.cn}
\author[add7,add6]{Soumyabrata~Dev\corref{mycorrespondingauthor}}

\address[add1]{The Fu Foundation School of Engineering and Applied Science, Columbia University}
\address[add3]{Biomedical Engineering Department, Pratt School of Engineering, Duke University}
\address[add7]{School of Computer Science, University College Dublin}
\address[add10]{Department of Electrical and Electronic Engineering, The University of Hong Kong}
\address[add2]{Robotics Institute, School of Computer Science, Carnegie Mellon University}
\address[add11]{Faculty of Science and Engineering, University of Nottingham Ningbo China}
\address[add4]{College of Intelligence and Computing, Tianjin University}
\address[add6]{The ADAPT SFI Research Centre}
\cortext[contrib]{Authors contributed equally to this research.}
\cortext[mycorrespondingauthor]{Corresponding author. Tel.: + 353 1896 1797.}
\ead{soumyabrata.dev@ucd.ie}


\begin{abstract}
Rainfall prediction remains a persistent challenge due to the highly nonlinear and complex nature of meteorological data. Existing approaches lack systematic utilization of grid search for optimal hyperparameter tuning, relying instead on heuristic or manual selection, frequently resulting in sub-optimal results. Additionally, these methods rarely incorporate newly constructed meteorological features such as differences between temperature and humidity, to capture critical weather dynamics. Furthermore, there is a lack of systematic evaluation of ensemble learning techniques and limited exploration of diverse advanced models introduced in the past one or two years. To address these limitations, we propose a  \uline{R}obust ensemble Le\uline{A}rning gr\uline{I}d search-tu\uline{N}ed fram\uline{E}wo\uline{R}k (RAINER) for rainfall prediction. RAINER incorporates a comprehensive feature engineering pipeline, including outlier removal, imputation of missing values, feature reconstruction, and dimensionality reduction via Principal Component Analysis (PCA). The framework integrates novel meteorological features to capture dynamic weather patterns and systematically evaluates non-learning mathematical-based methods and a variety of machine learning models, from weak classifiers (e.g., KNN, LASSO, and Random Forest) to advanced neural networks such as Kolmogorov-Arnold Networks (KAN). By leveraging grid search for hyperparameter tuning and ensemble voting techniques, RAINER achieves state-of-the-art results within real-world dataset. Extensive experiments demonstrate the framework’s superior performance across diverse metrics, such as Accuracy, Precision, Recall, F1-score, and AUC, highlighting the pivotal role of feature engineering, grid search, and ensemble learning in rainfall prediction tasks. 

\end{abstract}


\end{frontmatter}

\section{Introduction}
\label{sec:intro}
Geoscience has profound implications for tackling global challenges, spanning meteorology, climate resilience, disaster management, and sustainable urban planning. In recent years, data mining, machine learning, and deep learning have emerged as transformative tools in predicting patterns, enabling the extraction of valuable patterns from complex datasets and fostering advancements in various forecasting domains including healthcare prediction \cite{alaa2018survival, rajkomar2019healthcare, shan2021mlhealthcare, DEV2022100032}, where models assist in disease diagnosis and patient outcome prediction; entertainment trend prediction \cite{chen2023entertainment, WANG2024100601, xu2024aligngroup, xu2024mentor, xu2024fourierkan}, focusing on content popularity and audience preference analysis, enabling tailored recommendation systems; and geoscience forecasting \cite{gupta2023moistureforecast, li2023drought, huang2023earthquake, gupta2023glacierforecast}, which addresses critical environmental and geological challenges. Among the various geoscience prediction domains, multiple specific tasks in geoscience and remote sensing have greatly benefited from machine learning and deep learning-based methods, which have been applied to tasks such as land cover classification and vegetation mapping \cite{liu2019deep, schulz2018mlmethods, talukdar2020landcover, cui2024superpixel, cai1010, li2025cp2m}, cloud segmentation \cite{ma2019cloud, 10640450, li2025ddunetdualdynamicunet}, seismic activity prediction \cite{zhang2022seismicdl, huang2023earthquake, sarkar2020seismicml}, soil moisture prediction \cite{beck2019soilmoisture, yao2021dnnsoil, gupta2023moistureforecast}, drought assessment \cite{huang2022droughtdl, mishra2021drought, li2023drought}, glacier dynamics modeling \cite{lee2022glacierml, berger2023iceflow, gupta2023glacierforecast}, air temperature prediction \cite{9703606, cifuentes2020temperature, eshra2021comparison}, agriculture monitoring \cite{cui2024real}, and rainfall prediction \cite{hassan2023mlrainfall, shaharudin2022predictive, das2022rainfall, kumar2023overcastpredict}. Despite these advancements, each of these domains continues to present unique challenges. Remote sensing often involves the integration of heterogeneous data sources and the need for high-resolution imagery, which can be computationally intensive to process. Cloud segmentation requires robust models capable of handling noise and occlusions in satellite imagery. Rainfall prediction remains particularly challenging due to the inherent variability and nonlinearity of meteorological data, highlighting the importance of effective feature extraction and model generalization. These challenges underscore the need for continued research and innovation in the application of machine learning to geoscience problems.

Among these domains, rainfall prediction stands out due to its intricate dependence on multiple meteorological variables such as humidity, temperature, and atmospheric pressure. Feature extraction and dimensionality reduction techniques, including PCA, have been employed to simplify high-dimensional data while preserving key relationships \cite{pathan2022efficient, kustiyo2021analysis, meng2020dimension}. Additionally, using machine learning classifiers such as Random Forests, Gradient Boosting, Support Vector Machines (SVM), and XGBoost for rainfall prediction tasks has demonstrated superior performance compared to traditional statistical methods. These ML classifiers excel at capturing nonlinear relationships and handling high-dimensional datasets, enabling robust predictions and better generalization \cite{fan2018machine, sahin2020assessing, li2020rainfall}. However, existing approaches for rain prediction face several challenges below, and our motivation is to solve these issues.

\begin{itemize}
    \item  Some relevant work fails to fully leverage grid search for optimal hyperparameter tuning, only relying on heuristic selection. Furthermore, some studies that utilize grid search do not perform preliminary exploration to observe parameter trends, instead relying on arbitrary or heuristic parameter ranges, thereby increasing the likelihood of obtaining sub-optimal output. 
    \item Many existing models fail to leverage newly constructed features, such as the differences between temperature and humidity, which are critical for capturing specific meteorological dynamics.
    \item Ensemble methods, including voting and advanced integration techniques, have been insufficiently explored, particularly for assessing the combined effects of multiple models.
    \item Advanced models such as KAN \cite{liu2024kan}, have rarely been evaluated in comparison to methods like MLP, Transformer, LSTM, and weaker classifiers in this domain, limiting insights into their effectiveness in rainfall prediction.
\end{itemize}

To ameliorate the aforementioned problems, we propose a hybrid approach that integrates systematic data analysis with advanced modeling techniques to enhance rainfall prediction accuracy. Our approach includes a robust data preprocessing pipeline (e.g., imputing missing values, removing outliers, and dropping highly correlated features), feature construction, correlation analysis, and dimensionality reduction using PCA. Together, these steps form a comprehensive data processing system. Additionally, we conduct extensive experiments to explore parameter tuning and dataset partitioning strategies for each classifier, employing grid search to identify the relatively optimal parameters for a variety of weak and strong classifiers. We also evaluate the impact of ensemble methods such as voting on improving prediction performance. The main contributions of our RAINER are threefold:

\begin{itemize}
    \item We develop a robust and systematic data analysis and processing pipeline including feature construction (e.g., temperature/humidity max differences) and PCA on real-world data, which ensures high-quality inputs for predictive modeling.
    \item We perform extensive exploration on diverse non-learning mathematical and weak-to-advanced learning-based approaches, investigating optimal parameter settings through grid-search and examining the effects of ensemble voting methods.
    \item We conduct comprehensive experiments with detailed performance evaluation across multiple metrics (e.g., Accuracy, Precision, Recall, ROC Curve, AUC, and F1-score), providing quantitative comparisons of various methods and four distinct feature engineering strategies in rainfall prediction tasks, highlighting their respective strengths and limitations.
\end{itemize}


The structure of this paper is as follows. Section 2 reviews related work. Section 3 introduces the dataset and outlines the assumptions made for the experiments. Section 4 explains the data preprocessing and feature engineering process. Section 5 presents the methods, detailing non-learning mathematical approaches and weak/advanced learning-based models. Section 6 conducts experiments with quantitative analysis and ROC comparisons, and Section 7 concludes the paper with discussions on future work.

\section{Related Work}
\label{sec:relatedwork}

\subsection{Non-learning Approaches (e.g., mathematical, physical, statistical, and hybrid) for Rainfall Prediction}

Traditional approaches to rainfall prediction have predominantly relied on mathematical and statistical models. These models aim to describe atmospheric processes using physical equations or probabilistic frameworks. Mathematical models are rooted in the principles of physics and fluid dynamics, such as the Navier-Stokes equations, which are employed to simulate atmospheric motion and weather conditions. Numerical Weather Prediction (NWP) models are among the most prominent, integrating multiple physical equations to forecast rainfall over specific time horizons \cite{bauer2015numerical, steiner1995climatological}. Advanced formulations, such as those leveraging probabilistic mathematical structures, have been proposed to enhance rainfall prediction by accounting for variability across multiple temporal and spatial scales \cite{matricciani2011mathematical}. Additionally, spatiotemporal modeling techniques have been developed to enhance the precision of NWP by incorporating aerosol transport and other regional variables \cite{lee2024spatiotemporal}. While these models provide detailed simulations, their performance is often constrained by computational demands and the resolution of input data. Statistical methods, such as regression analysis, autoregressive integrated moving average (ARIMA), and Markov chain models, have also been widely used for rainfall prediction \cite{li2017markov, aksoy2018comparison}. These approaches rely on historical weather patterns and assume linear or probabilistic relationships between meteorological variables. However, they often fail to capture the nonlinear and chaotic nature of rainfall dynamics. For example, ARIMA models perform well in capturing short-term trends but struggle with long-term variability in rainfall sequences \cite{MP-NbCbFT7sJ}. An extension of ARIMA, known as Seasonal ARIMA (SARIMA), builds upon its predecessor by explicitly modeling seasonality in time series data, making it particularly effective in regions with pronounced seasonal variations. SARIMA has demonstrated its utility not only for capturing rainfall variability but also for addressing other seasonally driven phenomena where seasonal components serve as critical predictors \cite{inproceedings111}. For example, SARIMA models have been successfully applied to predict monsoonal rainfall patterns in tropical regions, improving both accuracy and interpretability compared to traditional ARIMA models \cite{gomez2024sarima}. Recent advances have combined physical and statistical methods to overcome individual limitations. Statistical downscaling models, which integrate large-scale atmospheric variables from physical models with local rainfall data, have shown promise in improving prediction accuracy for diverse geographical contexts \cite{wilby2004statistical, raje2010comparison}. Hybrid methods like these highlight the potential of blending domain knowledge with empirical relationships. For instance, Model Output Statistics (MOS) techniques have been found effective in refining precipitation forecasts by improving the output of global forecasting systems \cite{das2024feasibility}. Additionally, integrated mathematical-statistical frameworks have been applied in radar calibration for rainfall estimation, providing a more nuanced understanding of precipitation events \cite{4_-w0GzEho0J, wang2024deep}.

\subsection{Learning-based Methods for Rainfall Prediction}

Learning-based methods have emerged as transformative tools for rainfall prediction, addressing the limitations of traditional approaches by capturing nonlinear relationships and processing high-dimensional data. Supervised learning methods, including Random Forests (RF), SVM, and Gradient Boosting Machines (GBM), are commonly used for rainfall prediction. These models are capable of handling complex, multidimensional data while providing robust predictions. For instance, Random Forests excel at feature selection and generalization, while Gradient Boosting optimizes predictive performance by iteratively reducing errors \cite{sahour2021random, tao2022integration}. In addition, Bayesian approaches have been explored to enhance interpretability and probabilistic forecasting in ML frameworks \cite{MP-NbCbFT7sJ}. Deep learning models, such as Recurrent Neural Networks (RNNs) and Long Short-Term Memory (LSTM) networks, have been extensively studied for their ability to capture temporal dependencies in rainfall data \cite{wu2010prediction, pham2022evaluation}. Furthermore, recent advancements highlight the incorporation of spatial features, such as aerosol concentration and land-cover data, into deep learning frameworks to improve rainfall predictions \cite{lee2024spatiotemporal}. Hybrid machine learning techniques, such as hypertuned wavelet convolutional neural networks (WCNNs) combined with LSTMs, have shown improved accuracy in long-term rainfall forecasts and hydroelectric management \cite{stefenon2024hypertuned}. Convolutional Neural Networks (CNNs) are a widely recognized category of deep learning models, extensively utilized for analyzing and processing 2D image data~\cite{WANG2022102243, WANG2021, BATRA2022200039}. Specifically, CNNs-like encoders have been involved in multiple computer vision tasks including autonomous driving \cite{WANGCAR2022}, medical imaging (e.g., semantic segmentation, instance segmentation, and detection) \cite{tang2024optimized, pan2024accurate}, saliency object detection \cite{li2023daanet, Li_2024_BMVC}, human-robot interaction and perception \cite{zhenqi23, HuoAtten, wang2024airshot}, as well as object manipulation \cite{zhu2023fanuc, wang2024onls, Lin2024JointPT, Huo2023HumanorientedRL}. In recent years, CNNs have also been applied to rainfall prediction tasks, particularly for analyzing spatial data from satellite imagery \cite{tao2022integration}. These methods significantly outperform traditional statistical techniques by capturing the nonlinear and chaotic nature of rainfall dynamics. Additionally, innovative hybrid approaches integrating GNSS-based atmospheric data with machine learning models have demonstrated significant potential in enhancing rainfall prediction accuracy \cite{haji2024mlgnss}. Hybrid deep learning architectures, such as those combining LSTMs with attention mechanisms, are gaining traction for their ability to enhance long-sequence forecasting \cite{liu2024kan}. Ensemble methods, such as bagging, boosting, and stacking, have gained popularity for their ability to combine predictions from base models. Techniques like XGBoost and voting improve model robustness and reduce overfitting \cite{liu2024kan, sahour2021random}.

\section{Dataset}
The dataset is provided by the Australian Bureau of Meteorology (BoM) and contains 145,460 records with 23 variables describing meteorological features and observation dates. \footnote{\url{http://www.bom.gov.au/climate/data}} The data spans 10 years of daily weather observations from multiple weather stations across Australia. Key variables include date and location for temporal and spatial references, meteorological features such as temperature, rainfall, wind, humidity, pressure, cloud cover, and sunshine, as well as binary indicators RainToday and RainTomorrow for rainfall prediction. The goal of this study is to predict the target variable RainTomorrow by using these features.\footnote{\url{http://www.bom.gov.au/climate/dwo/}}

\section{Data Precessing \& Feature Engineering}
In our proposed RAINER framework, we performed data cleaning to address missing values, outliers, and inconsistencies in the original dataset. Variables with excessive missingness were removed while missing data in other features were imputed using statistical methods. Outliers in features such as Rainfall and WindGustSpeed were capped, and numerical variables were normalized to ensure consistency. In the feature extraction process, we deleted or merged redundant features and constructed new ones, such as creating the max difference of MaxTemp and MinTemp and dropping the original MaxTemp and MinTemp, as well as creating the max difference of Humidity9am and Humidity3pm and dropping the original Humidity9am and Humidity3pm. In addition, we applied PCA and t-SNE methods to facilitate dimensionality reduction and cluster analysis, which helped validate the effectiveness of feature selection and the rationality of the newly constructed features.

\subsection{Data Preprecessing}
\subsubsection{Handling missing data}
\begin{figure}[h]
	\centering
	\includegraphics[width=0.9\textwidth]{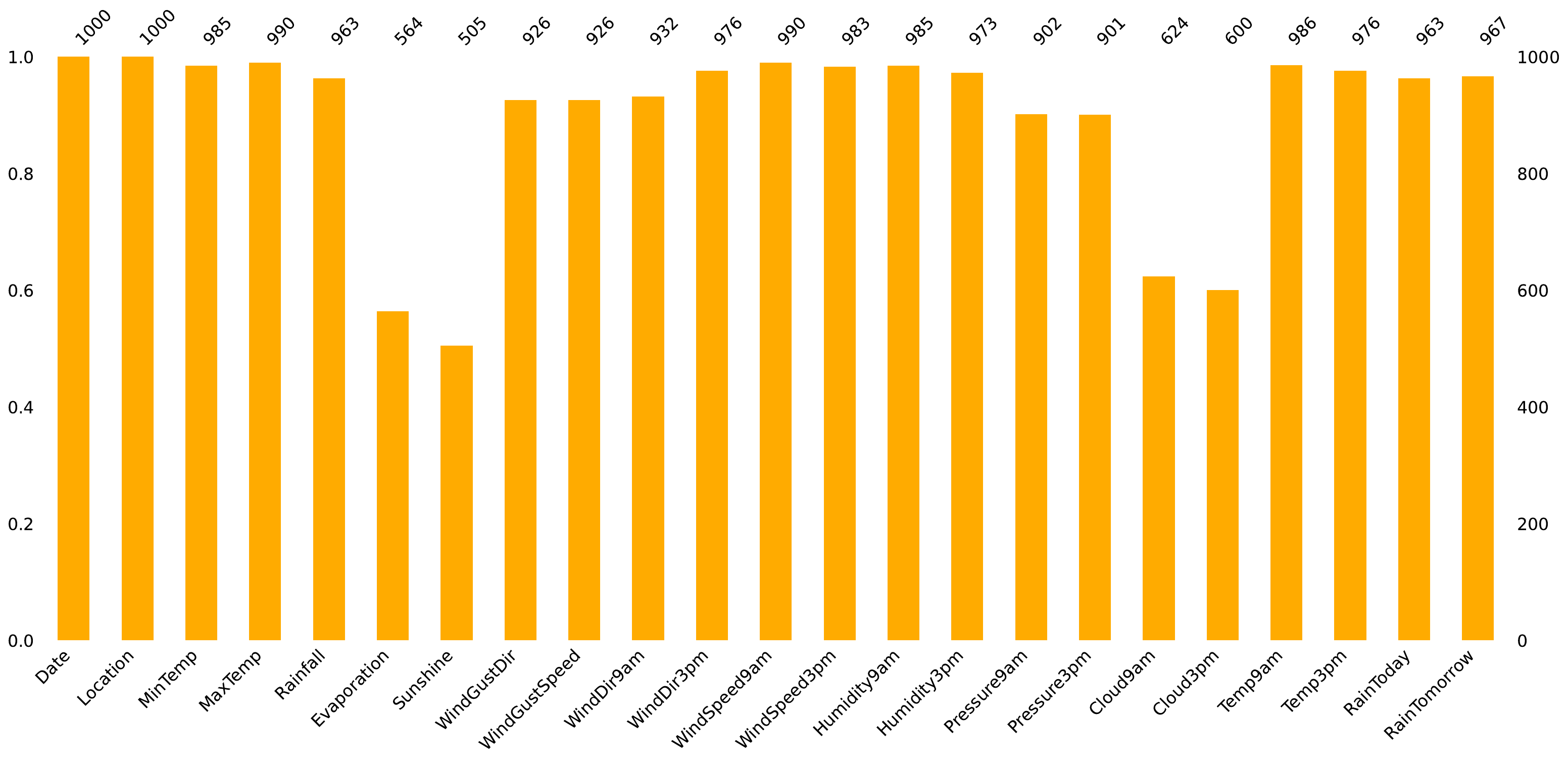}
	\caption{Bar Chart of missing values for meteorological features. Features such as "Evaporation" and "Cloud9am" show the highest missingness, while others like "Location" and "MinTemp" have near-complete data. High-missingness features were excluded to ensure unbiased analysis.}
        \label{fig:missing_bar}
\end{figure}

Handling missing data is a critical step in data preprocessing, as it directly impacts the reliability and accuracy of subsequent analyses. In this study, we addressed missing values by identifying high-missingness features, categorizing remaining features, and applying imputation strategies. To further quantify the extent of missingness, Figure~\ref{fig:missing_bar} was generated to visualize the proportion of missing values for each feature. Based on this analysis, we determined that these features were unsuitable for reliable analysis due to their high missingness and limited relevance to the target variable. Consequently, these features were removed to simplify subsequent processing and avoid introducing bias.

After removing high-missingness features, 19 features remained in the dataset. These features were categorized into two groups based on their data types and missingness levels:

\begin{itemize}
    \item \textbf{Numerical Features}: Features such as Rainfall, Temp9am, and Pressure3pm exhibited moderate missingness (less than 10\%). Mean imputation was applied to preserve their central tendencies while ensuring minimal distortion of their distributions.
    \item \textbf{Categorical Features}: Features such as RainToday and WindGustDir had low missingness and were imputed using the mode to retain the most frequent categories.
\end{itemize}

\begin{figure}[h]
	\centering
        \includegraphics[width=0.9\textwidth]{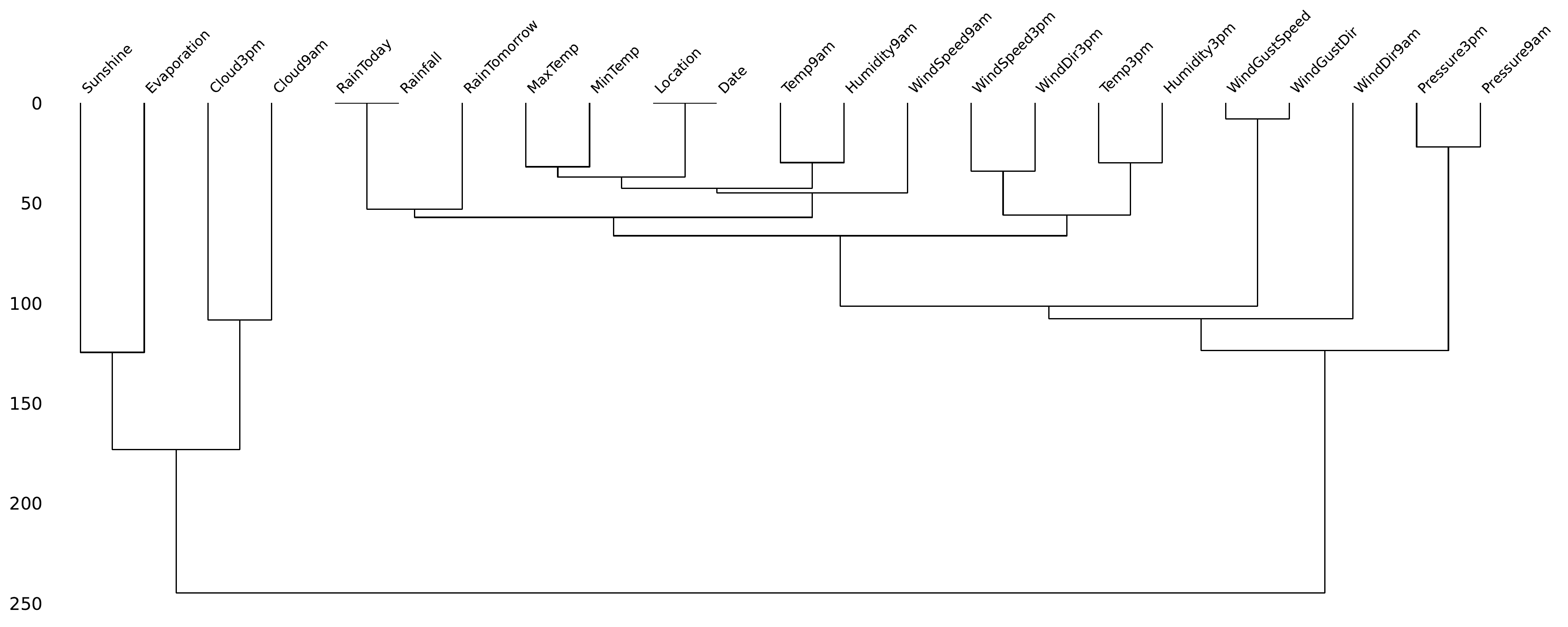}
        \caption{Weather attribute dendrogram. It highlights relationships such as the strong clustering of "Pressure9am" and "Pressure3pm.}
        \label{fig:dendrogram_missing}
\end{figure}

\begin{figure}[h]
	\centering
        \includegraphics[width=0.8\textwidth]{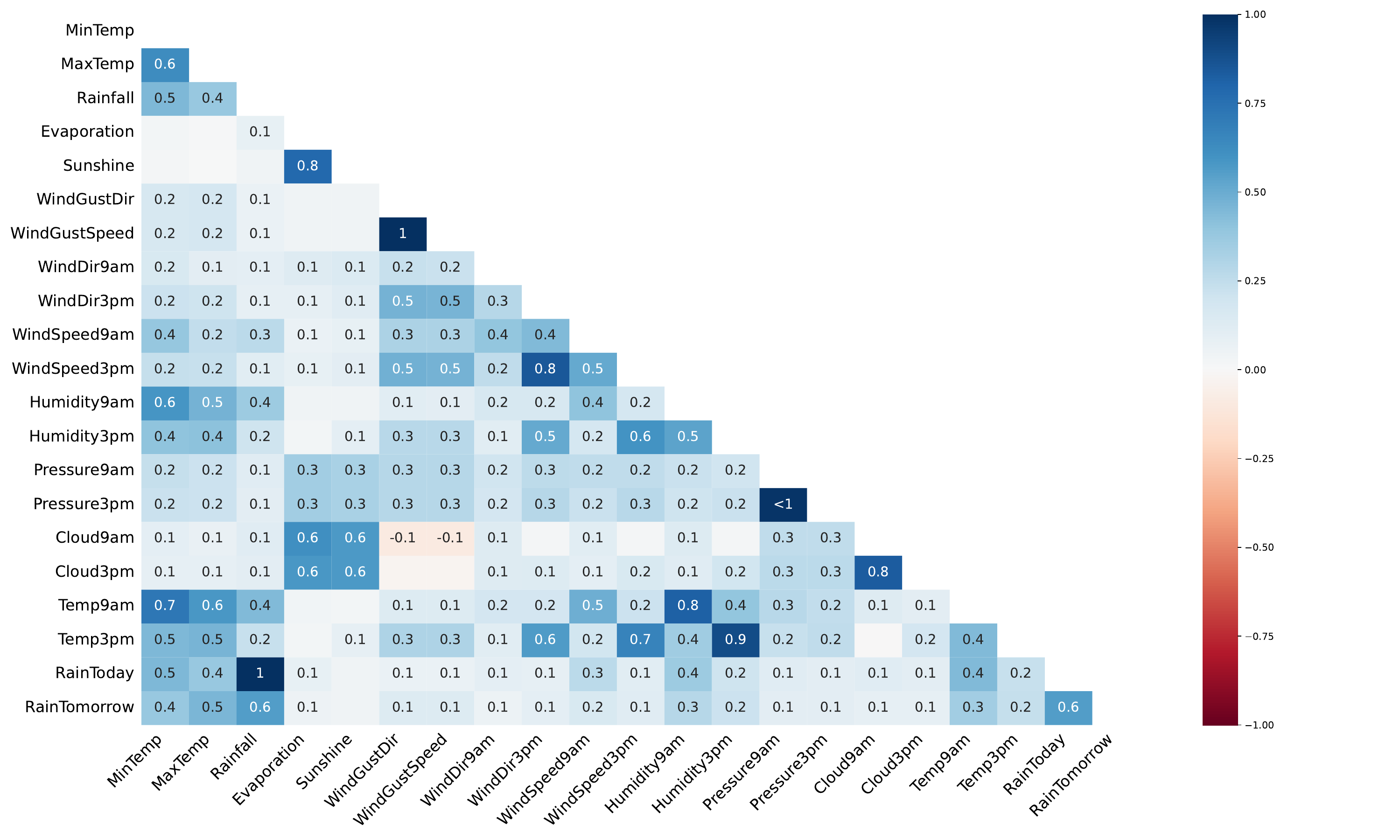}
        \caption{Weather feature correlation matrix. The correlation heatmap illustrates the dependencies between features, revealing strong correlations such as those between "Humidity9am" and "Humidity3pm".}
        \label{fig:correlation_heatmap}
\end{figure}

As Figure~\ref{fig:dendrogram_missing} shows, a dendrogram was generated to cluster features based on the similarity of their missing data patterns. This clustering highlighted relationships, such as the similarity between Pressure9am and Pressure3pm, which informed our imputation strategy for these features. Additionally, as Figure \ref{fig:correlation_heatmap} shows, a correlation heatmap was used to examine feature relationships, ensuring that the selected imputation methods aligned with observed dependencies.

\begin{figure}[h]
    \centering
    \includegraphics[width=1\textwidth]{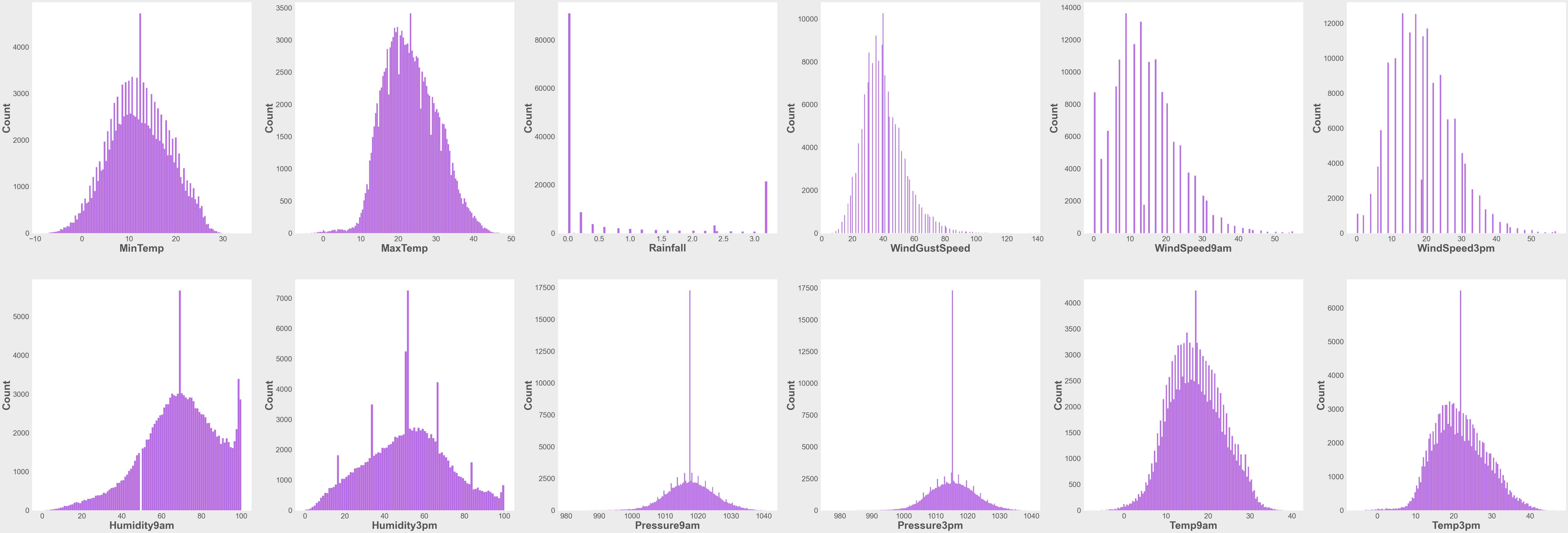}
    \caption{Histograms comparing feature distributions before and after imputation. The x-axis represents various meteorological attributes and the y-axis indicates the count of observations for each attribute. These histograms demonstrate that the imputation methods preserved the central tendencies and overall distributions of the data.}
    \label{fig:histogram_imputation}
\end{figure}

 As Figure~\ref{fig:histogram_imputation} shows, the impact of our imputation methods on numerical features was assessed using histograms, which compared the distributions of selected features before and after imputation. These histograms confirmed that the central tendencies and overall distributions of features were preserved, validating the effectiveness of our imputation methods. 

\begin{figure*}[h]
    \centering
    \begin{minipage}[b]{0.49\textwidth}
        \centering
        \includegraphics[width=\textwidth]{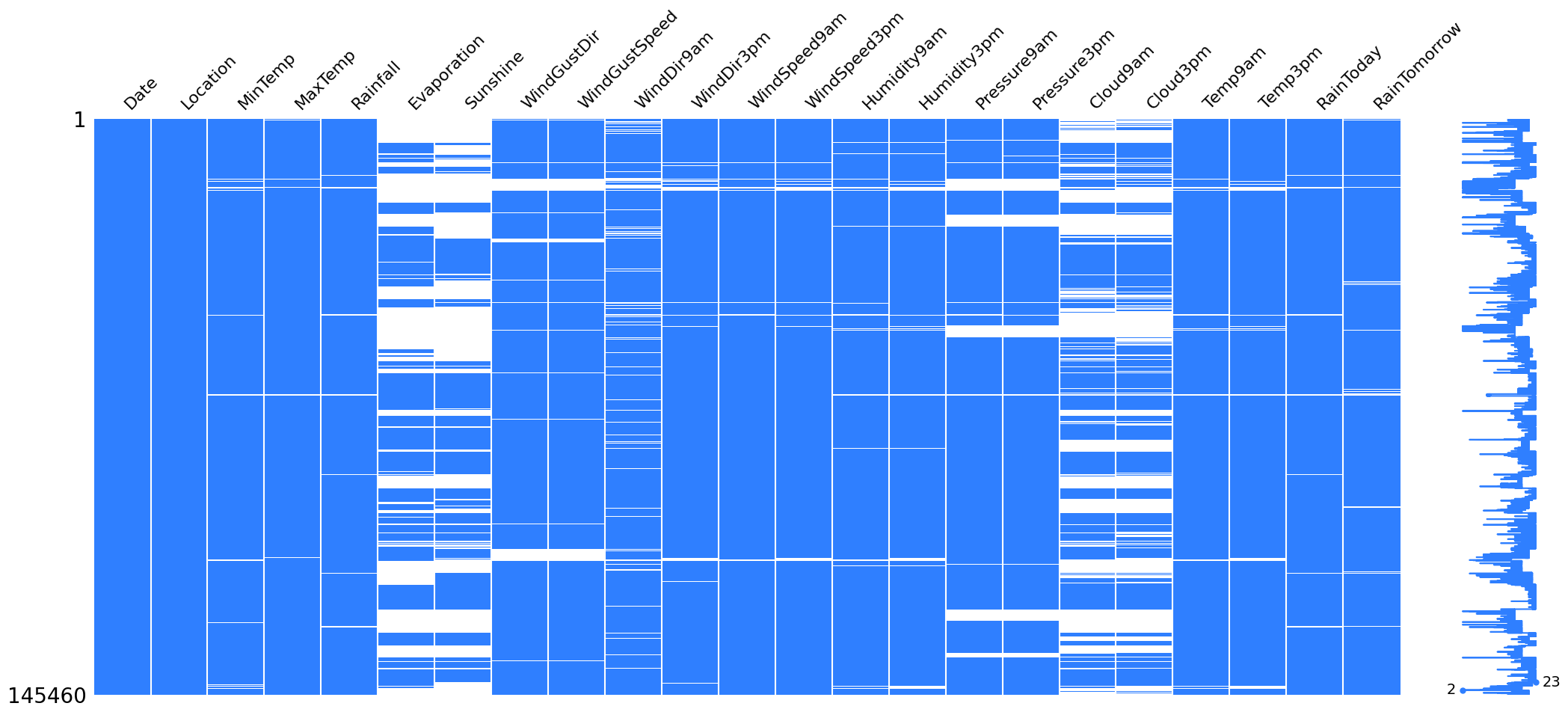}
        \caption{Missing value matrix. The missing value matrix displays the distribution of missing values across different features.}
        \label{fig:missing_matrix_before}
    \end{minipage}
    \hfill
    \begin{minipage}[b]{0.49\textwidth}
        \centering
        \includegraphics[width=\textwidth]{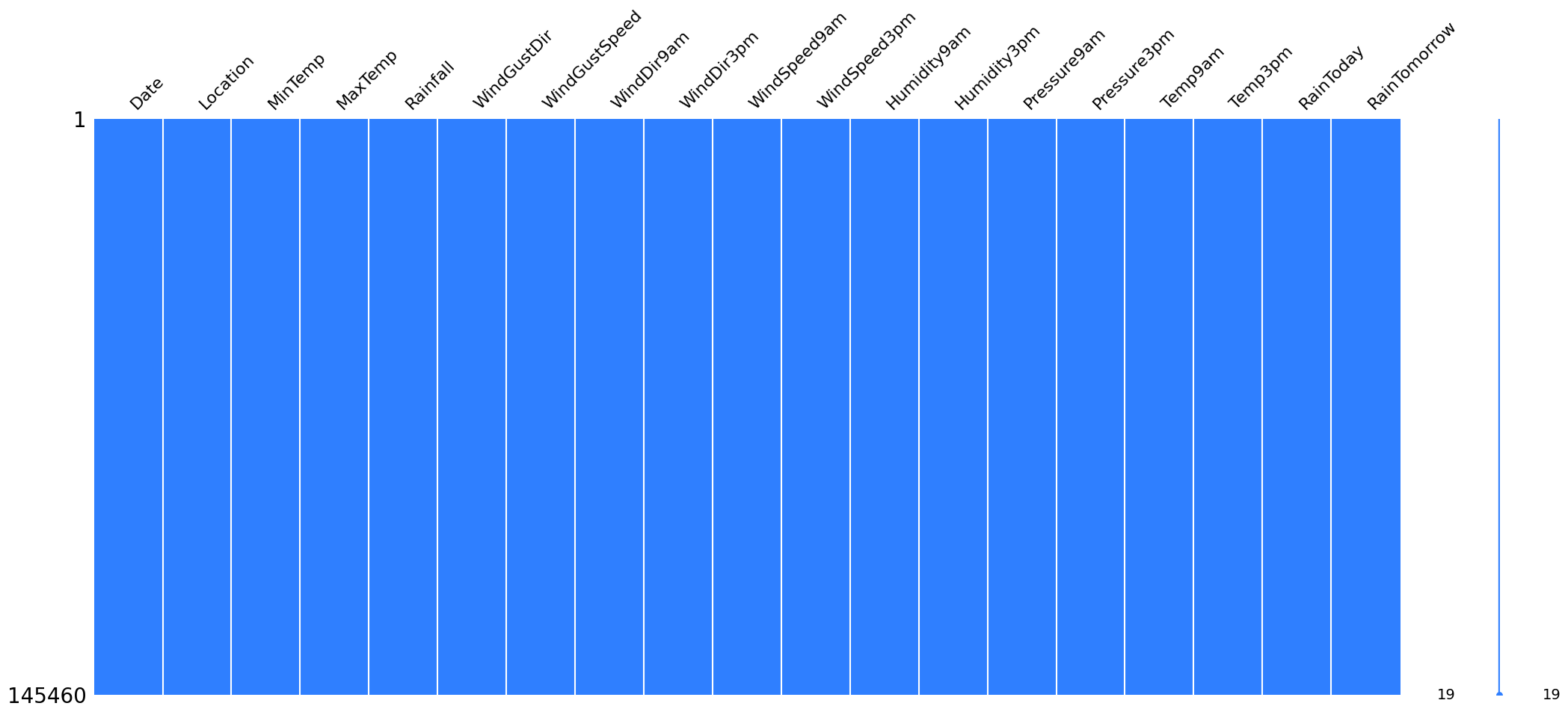}
        \caption{Post-imputation missing value matrix. The post-imputation missing value matrix confirms the resolution of all missing data.}
        \label{fig:missing_matrix_after}
    \end{minipage}
\end{figure*}

As shown in Figure~\ref{fig:missing_matrix_before}, the initial missing value matrix revealed significant gaps in certain features, including Evaporation, Sunshine, Cloud9am, and Cloud3pm, which exhibited missingness rates exceeding 30\%. Finally, as Figure~\ref{fig:missing_matrix_after} shows, the post-imputation missing value matrix confirmed the resolution of all missing entries. The absence of white spaces in this matrix demonstrated that the dataset is now complete, providing a consistent foundation for further analysis and modeling.

\subsubsection{Outlier Detection \& Treatment}

\begin{figure}[h]
	\centering
        \includegraphics[width=\textwidth]{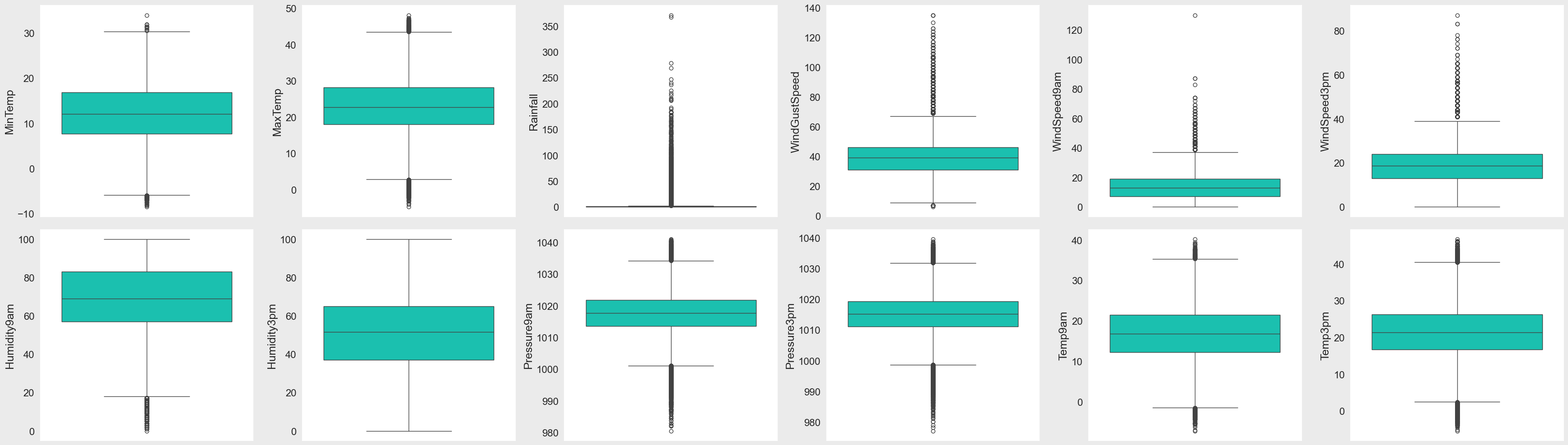}
        \caption{Boxplots of numerical features before outlier handling.}
        \label{fig:boxplot_before}
\end{figure}

Outlier detection and treatment is an essential part of data preprocessing, as extreme values can distort statistical analyses and negatively impact model performance. In this study, we began by visualizing numerical features using boxplots in Figure~\ref{fig:boxplot_before} to identify outliers. Features such as Rainfall, WindSpeed9am, and WindSpeed3pm exhibited extreme values well beyond their typical ranges. For instance, the Rainfall feature displayed values higher than expected, which could affect the reliability of downstream analyses.

\begin{figure}[h]
	\centering
        \includegraphics[width=\textwidth]{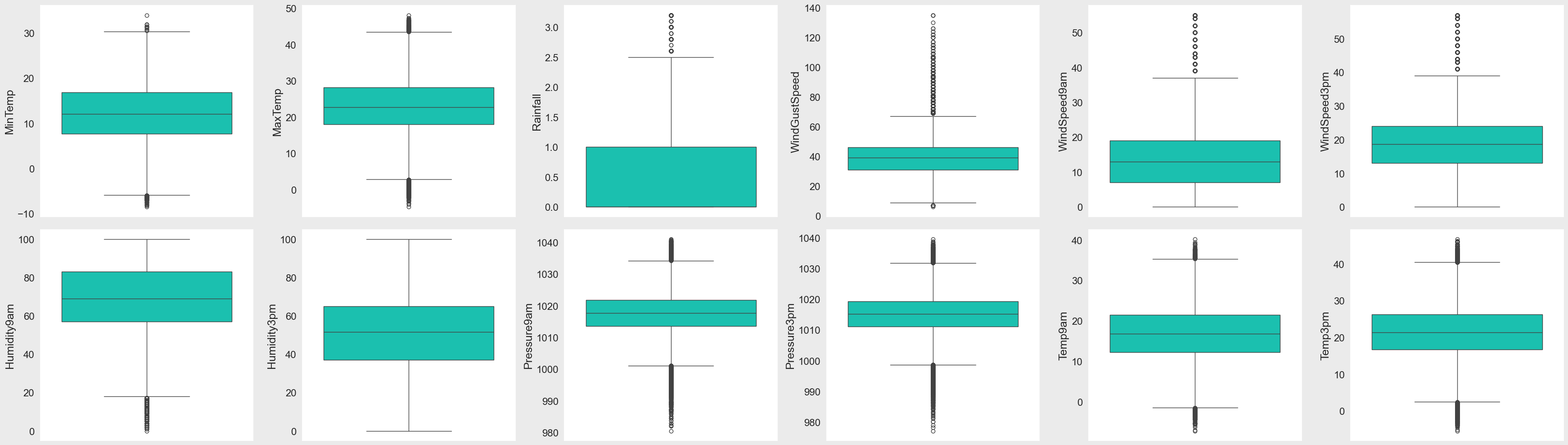}
        \caption{Boxplots of numerical features after outlier handling.}
        \label{fig:boxplot_after}
\end{figure}

To address these outliers, we adopted a capping strategy, replacing extreme values exceeding predefined thresholds with the thresholds themselves. Thresholds were determined based on domain knowledge and visual inspection of the data. Specifically, Rainfall values above 3.2 mm were capped at 3.2 mm, while WindSpeed9am and WindSpeed3pm values exceeding 55 km/h and 57 km/h, respectively, were adjusted to these thresholds. Post-treatment boxplots as Figure~\ref{fig:boxplot_after} shows, confirmed the effectiveness of this approach, as the adjusted distributions no longer contained extreme values while preserving their central tendencies. This ensured that the dataset remained robust and reliable for subsequent analysis and modeling.

\subsection{Feature Selection \& Reconstruction}
\begin{figure}[h] 
        \centering 
        \includegraphics[width=0.75\textwidth]{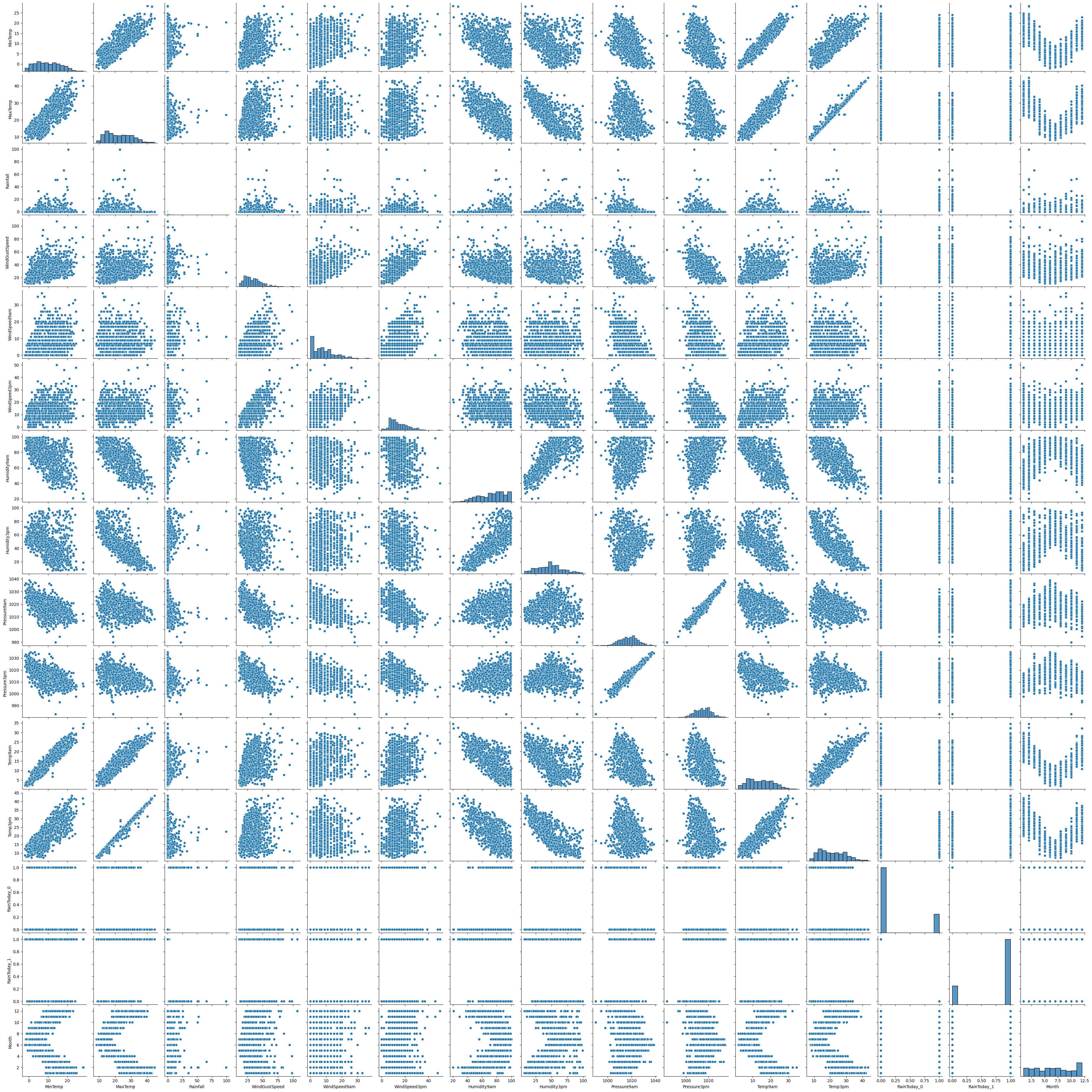} 
        \caption{Pair plot of features. The pair plot visualizes pairwise relationships among the first 950 records of selected features, providing insights into feature correlations, distributions, and potential patterns.} 
        \label{fig 2.9}
\end{figure}

\begin{figure}[h] 
        \centering 
        \includegraphics[width=0.9\textwidth]{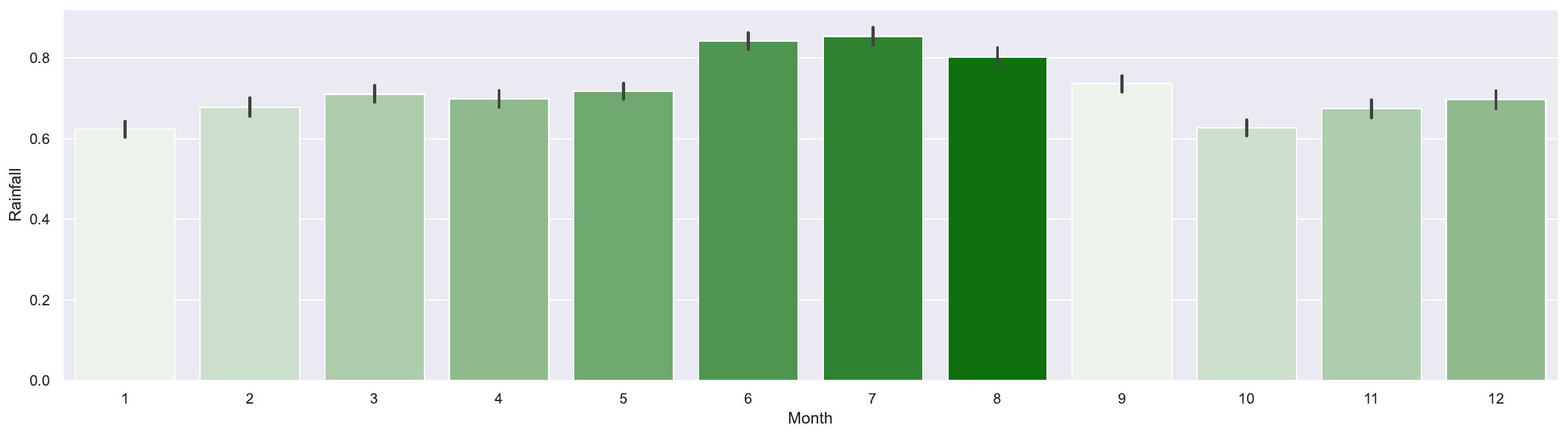} 
        \caption{Bar chart of average rainfall by month. The bar chart shows the average monthly rainfall, highlighting significant seasonal variations. June, July, and August experience the highest rainfall.} 
        \label{fig 2.10}
\end{figure}

To enhance the efficiency of data analysis and modeling, we performed several preprocessing steps and visualized the feature relationships using a pair plot below. The Date feature was converted to a DateTime format, and the Month was extracted as a new feature to facilitate time-based analysis. For categorical feature encoding, we used BinaryEncoder for the RainToday feature, which is a binary classification variable. Other categorical features such as Location, RainTomorrow, WindDir3pm, and WindGustDir were encoded using LabelEncoder to convert them into label classes suitable for modeling. After outlier handling, Figure \ref{fig 2.9} illustrates the pairwise relationships among the first 950 records. The pair plot provides insights into feature correlations and distributions, guiding further analysis. To address multicollinearity issues, we need to drop highly correlated features, which can improve model robustness and performance. Besides, we investigated rainfall trends across different months by creating a bar plot in Figure \ref{fig 2.10} depicting the relationship between the month and the average rainfall. The analysis shows that June, July, and August are peak rainfall seasons. 
The significant seasonal variation in rainfall supports further time series analysis to predict future rainfall trends.

\begin{figure}[h] 
        \centering 
        \includegraphics[width=0.9\textwidth]{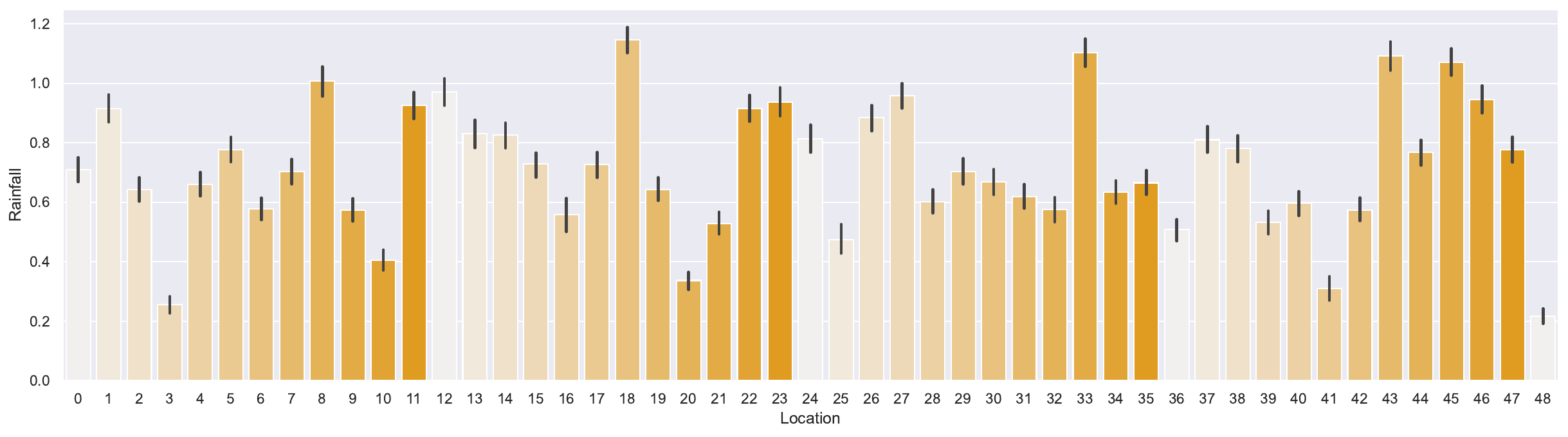} 
        \caption{Bar chart of average rainfall by location. The bar chart illustrates the average rainfall for different locations. Locations 18, 33, and 43 exhibit significantly higher rainfall compared to others.} 
        \label{fig 2.11} 
\end{figure}
\begin{figure}[h] \centering \includegraphics[width=\textwidth]{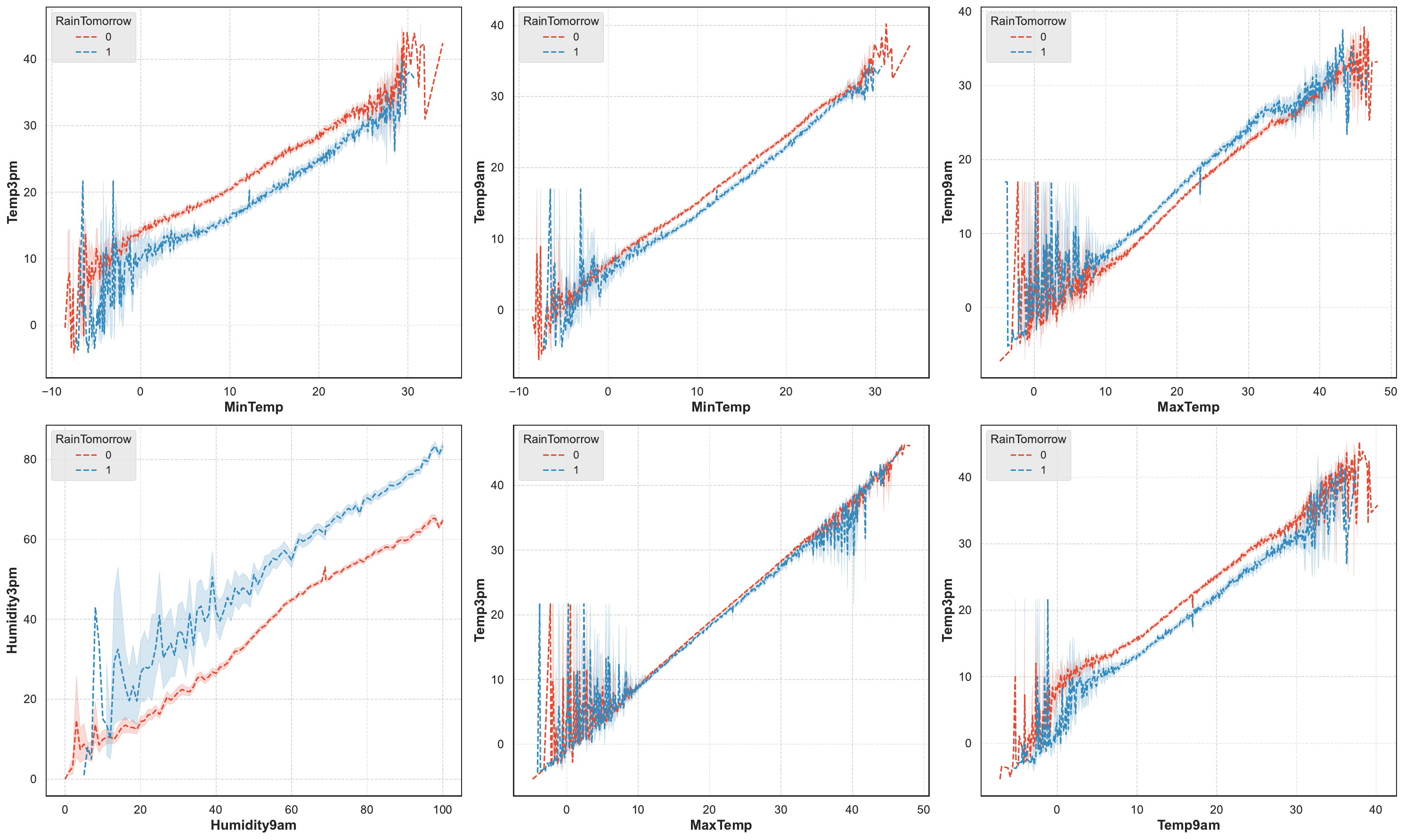} \caption{Line plots of relationships between meteorological features. The line plots depict the relationships between various meteorological features and the likelihood of rainfall the next day, differentiated by the "RainTomorrow" categories.} \label{fig 2.12} \end{figure}

To explore rainfall variations across different locations, we created a bar plot in Figure \ref{fig 2.11} illustrating the average rainfall for each location. The analysis indicates that cities numbered 18, 33, and 43 have significantly higher rainfall compared to other locations. These areas might experience higher rainfall due to geographical or climatic factors. The clear differences in rainfall between cities highlight location as a critical factor influencing rainfall, helping identify regions prone to heavy rainfall. We further explored the relationships between various meteorological features and how they relate to the likelihood of rainfall the next day. Figure \ref{fig 2.12} presents line plots of different feature pairs, with colors distinguishing the RainTomorrow categories. There is a strong linear relationship between MinTemp, and both Temp3pm and Temp9am can be modeled as shown in Equation~\eqref{eq:linear_model} as follows:

\begin{equation}
y = \beta_0 + \beta_1 x + \epsilon
\label{eq:linear_model}
\end{equation}where \(x\) represents MinTemp, \(y\) denotes Temp3pm or Temp9am, \(\beta_0\) and \(\beta_1\) are regression coefficients, and \(\epsilon\) is the error term. Regardless of whether it rains the next day, temperature features exhibit consistent trends.

The relationship between Humidity9am and Humidity3pm suggests that higher humidity levels are associated with an increased likelihood of rain the next day. This relationship is indicated by the conditional probability as shown in Equation~\eqref{eq:conditional_probability} as below:

\begin{equation}
P(\text{RainTomorrow} \mid X) = \frac{P(X \mid \text{RainTomorrow}) P(\text{RainTomorrow})}{P(X)}
\label{eq:conditional_probability}
\end{equation}where \(X\) represents observed meteorological features, \(P(\text{RainTomorrow} \mid X)\) is the posterior probability of rain given \(X\), \(P(X \mid \text{RainTomorrow})\) is the likelihood, \(P(\text{RainTomorrow})\) is the prior probability of rain, and \(P(X)\) is the marginal probability of the features. This is further illustrated by the significant upward trend of the blue line in the plot.

Additionally, the relationships between MaxTemp and both Temp3pm and Temp9am as shown in Equation~\eqref{eq:dataset_representation} as following:

\begin{equation}
D = \bigcup_{i=1}^{n} \{(x_i, y_i, c_i)\}
\label{eq:dataset_representation}
\end{equation}where \(x_i\) and \(y_i\) represent paired feature values, \(c_i\) indicates the corresponding RainTomorrow category, and \(n\) is the number of observations. These relationships highlight clear temperature patterns grouped by RainTomorrow categories, providing valuable insights for understanding and predicting weather conditions.

\begin{figure}[h] 
        \centering 
        \includegraphics[width=0.8\textwidth]{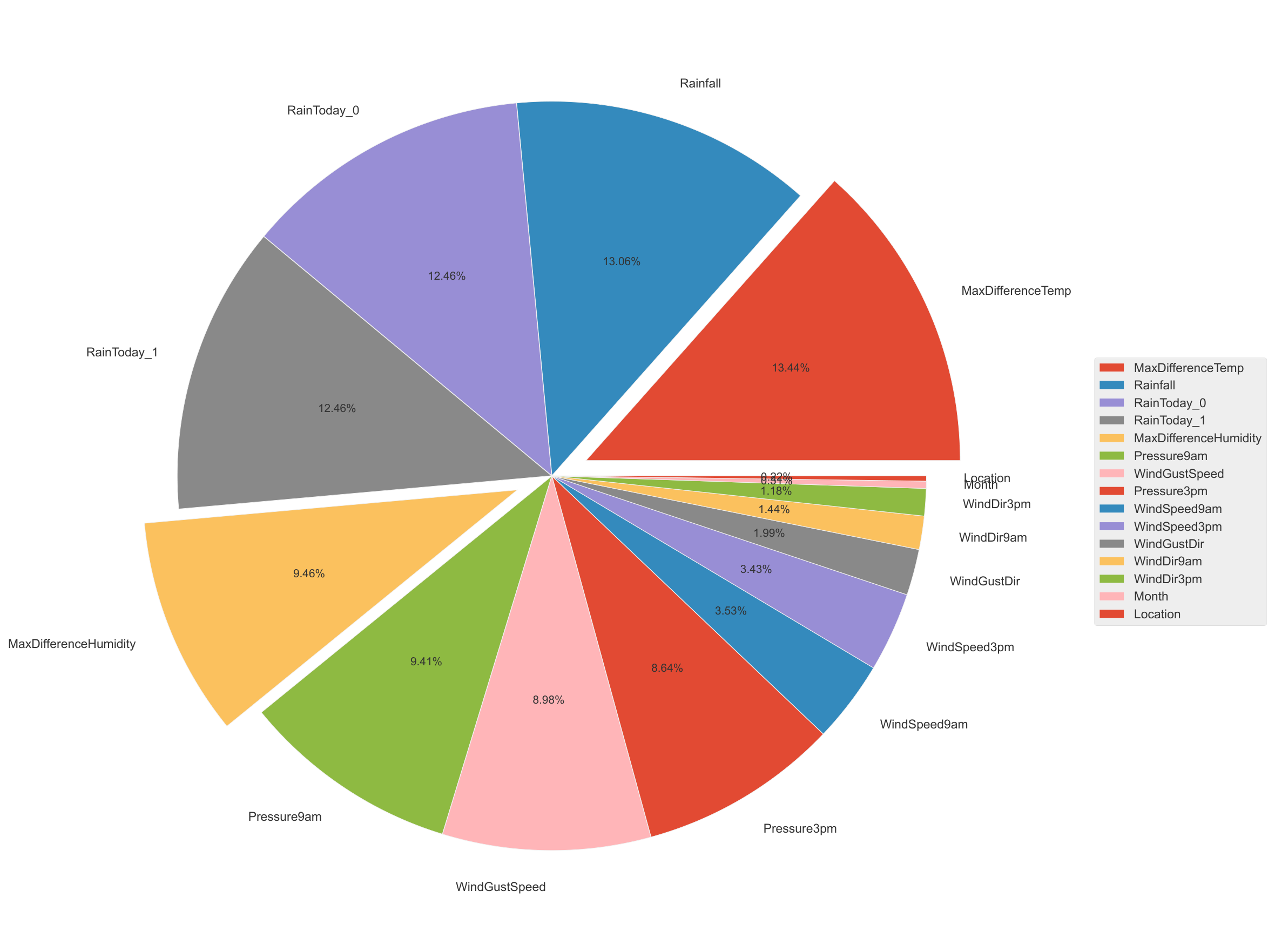} 
        \caption{Pie chart of feature correlations with "RainTomorrow". The pie chart illustrates the correlation weights of different meteorological features with "RainTomorrow." "MaxDifferenceTemp" and "Rainfall" are the most correlated features, accounting for 13.44\% and 13.06\% of the total correlation weight, respectively. Other significant features, including "RainToday," humidity, pressure, and wind speed, also contribute notably.} \label{fig 2.13} 
\end{figure}

Based on the above analysis, we concluded that specific meteorological features, such as temperature differences, rainfall, and humidity, play a significant role in predicting the likelihood of rain. To further evaluate the correlation between various features and the likelihood of rain the next day, we created a pie chart in Figure \ref{fig 2.13} that depicts the correlation weights of different features. The analysis shows that MaxDifferenceTemp and Rainfall are the most correlated features with RainTomorrow, accounting for 13.44\% and 13.06\% of the correlation weight, respectively. This supports the decisions made in our feature selection and reconstruction process. Features like RainToday, humidity, pressure, and wind speed also show high correlation, indicating their significant impact on predicting rainfall. The pie chart validates the effectiveness of our chosen features and processing methods, providing a strong foundation for subsequent model training.

\begin{figure}[h] 
        \centering 
        \includegraphics[width=0.8\textwidth]{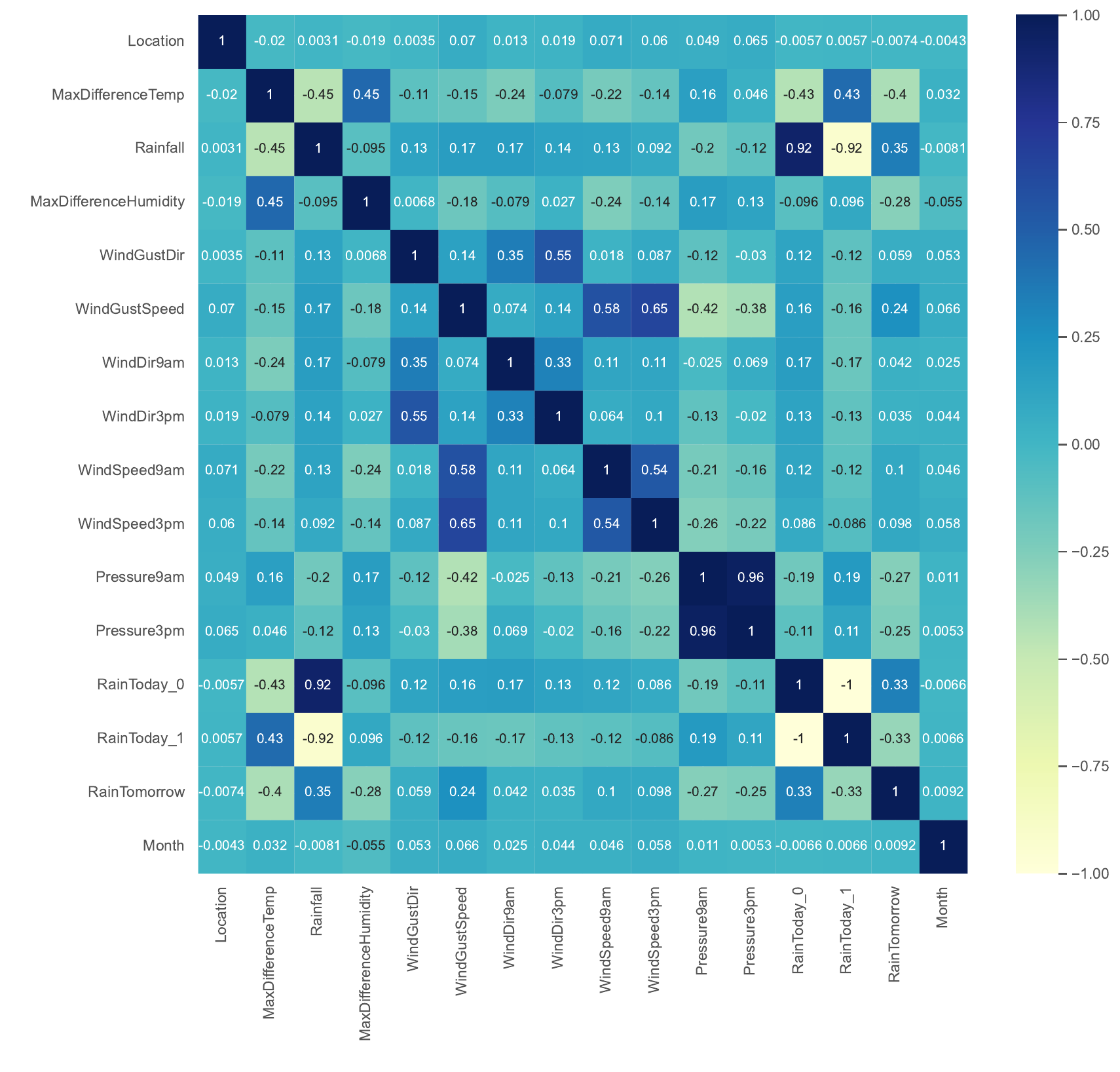} 
        \caption{Correlation heatmap of features. The heatmap visualizes the strength of correlations between features, with darker colors representing stronger correlations and lighter colors indicating weaker ones. A strong positive correlation is observed between "Pressure9am" and "Pressure3pm",
        while the negative correlation between "RainToday\_0" and "RainToday\_1" reflects their binary encoding. Features like "Location" and "Month" show relatively low correlations.} 
        \label{fig 2.14} 
\end{figure}

After summarizing the features, we proceeded to balance the final dataset to ensure equitable representation of all categories, following preprocessing and the implementation of upsampling techniques \cite{HENDERSON2025100529, lemaitre2017imbalanced}. To gain a deeper understanding of the relationships between different features, we generated a correlation heatmap, as shown in Figure \ref{fig 2.14}. The color intensity represents the strength of the correlation, with darker colors indicating stronger correlations and lighter colors indicating weaker ones. The strongest correlation is observed between Pressure9am and Pressure3pm, close to 1, indicating consistent variations. Additionally, the negative correlation between RainToday\_0 and RainToday\_1 is expected due to their encoding. Features like Location and Month show relatively low correlations with most other features, suggesting they provide unique information that enhances model diversity. The heatmap confirms the effective scaling and adjustment of features, ensuring the dataset is well-prepared for subsequent analysis and modeling.

\subsection{Cluster Analysis}

After feature processing, we applied cluster analysis to validate whether the processed data clearly reflects category distributions. To achieve this, we utilized the t-distributed stochastic neighbor embedding (t-SNE) method, a widely used technique for dimensionality reduction \cite{Maaten2008}. The t-SNE process operates in two distinct phases. In the first phase, a probability distribution is constructed over pairs of data points in the high-dimensional space. Points that are similar are assigned higher probabilities, while dissimilar points are assigned lower probabilities. The second phase establishes a corresponding probability distribution in the low-dimensional space, and the algorithm minimizes the Kullback-Leibler (KL) divergence between these distributions by iteratively adjusting the positions of the points. Let $M$ represent the number of high-dimensional data points $x_1, \dots, x_M$. The t-SNE algorithm begins by calculating the pairwise similarities $q_{ab}$ between points $x_a$ $x_b$ in the high-dimensional space. For $a \neq b$, the similarity is defined as Equation~\eqref{eq:similarity_high}:

\begin{equation}
q_{ab} = \frac{\exp\left(-\frac{\lVert x_a - x_b \rVert^2}{2\sigma_a^2}\right)}{\sum_{l \neq a} \exp\left(-\frac{\lVert x_a - x_l \rVert^2}{2\sigma_a^2}\right)}
\label{eq:similarity_high}
\end{equation}where $\lVert x_a - x_b \rVert^2$ represents the squared Euclidean distance between $x_a$ and $x_b$, and $\sigma_a$ is the bandwidth parameter of the Gaussian kernel for point $x_a$. Additionally, we assign $q_{aa} = 0$. It is important to note that the probabilities $q_{ab}$ are normalized, such that $\sum_b q_{ab} = 1$ for all $a$.

Next, the joint probability $q_{ab}$ is symmetrized as Equation~\eqref{eq:joint_probability} below:

\begin{equation}
q_{ab} = \frac{q_{b|a} + q_{a|b}}{2M}
\label{eq:joint_probability}
\end{equation}where $q_{b|a}$ and $q_{a|b}$ represent the conditional probabilities between points $x_a$ and $x_b$, and $M$ is the total number of data points. This ensures that $q_{ab} = q_{ba}$, $q_{aa} = 0$, and $\sum_{a,b} q_{ab} = 1$.

In the low-dimensional space, the similarity \( r_{ab} \) between points \( y_a \) and \( y_b \) is defined using a heavy-tailed Student's t-distribution, as shown below:

\begin{equation}
r_{ab} = \frac{\left(1 + \lVert y_a - y_b \rVert^2\right)^{-1}}{\sum_{m \neq n} \left(1 + \lVert y_m - y_n \rVert^2\right)^{-1}}
\label{eq:similarity_low}
\end{equation}where $\lVert y_a - y_b \rVert^2$ is the squared Euclidean distance between $y_a$ and $y_b$ in the low-dimensional space. This choice of distribution allows dissimilar points to be placed farther apart, enhancing cluster separability.

The primary objective of t-SNE is to minimize the KL divergence between the high-dimensional and low-dimensional distributions as Equation~\eqref{eq:kl_divergence} below:

\begin{equation}
\text{KL}(P \| Q) = \sum_{a \neq b} q_{ab} \log \frac{q_{ab}}{r_{ab}}
\label{eq:kl_divergence}
\end{equation}where $P$ and $Q$ represent the high-dimensional and low-dimensional distributions, respectively. The optimization is performed using gradient descent, iteratively adjusting the positions of $y_a$ in the low-dimensional space.

\begin{figure}[h]
    \centering
    \includegraphics[width=0.7\textwidth]{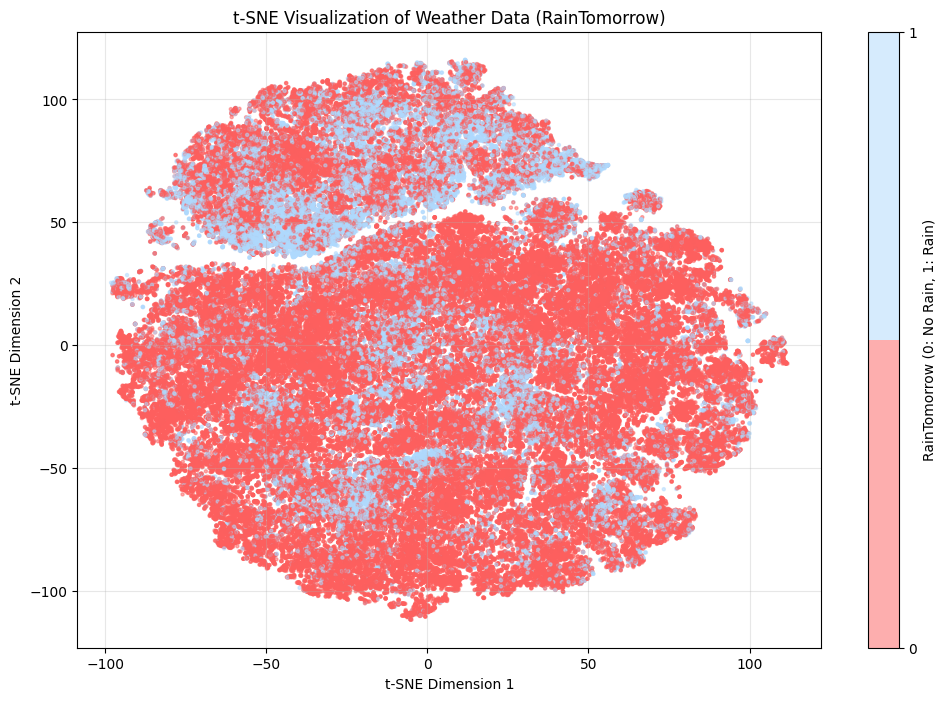}
    \caption{t-SNE visualization of weather data. The t-SNE plot shows distinct clusters for the target label categories "0" (no rain) and "1" (rain), confirming the effectiveness of applied feature processing/selection/construction strategy in our feature engineering pipeline.}
    \label{fig:tsne_visualization}
\end{figure}

At this stage, the two-dimensional representation of the processed weather data is obtained. Figure \ref{fig:tsne_visualization} illustrates the t-SNE results, showing distinct clusters for categories `0' (no rain) and `1' (rain). This visualization confirms that the applied feature processing and selection techniques effectively separate the two categories in the reduced-dimensional space. The results demonstrated that categories `0' and `1' were distinctly visible in the reduced-dimensional space, further confirming the effectiveness of our feature selection. 

\subsection{Principal Component Analysis}
In this study, we examine the variance within the dataset by utilizing Principal Component Analysis (PCA) \cite{Abdi2010}. PCA is a multivariate technique that transforms the original dataset into a series of linearly uncorrelated variables, known as principal components, which capture the maximum variance from the data. These components serve as compact summaries of the dataset's features. Although these new basis functions are not directly interpretable in physical terms, they are constructed as linear combinations of the original feature vectors. Instead of relying exclusively on conventional feature elimination strategies, we use dimensionality reduction to map the high-dimensional feature space onto a 2D representation, facilitating a deeper understanding of the relationships among features. PCA is particularly advantageous for predictive modeling when a few principal components explain the majority of the variance in the data. In this study, PCA is applied to analyze the 17-dimensional meteorological feature space by projecting it into a lower-dimensional subspace shown in Figure \ref{fig:biplots}. Furthermore, a concise explanation and mathematical foundations of PCA are shown below.

To formalize this, let $\mathbf{X}$ represent the variable matrix of dimensions $m \times n$, where $m$ denotes the number of meteorological features, and $n$ represents the total number of observations in the dataset. For this analysis, we have $m = 17$ and $n = 200,000$, corresponding to 17 meteorological features and 200,000 daily records. Each feature, $\mathbf{f}_j$, where $j = 1, 2, \dots, 17$, is extracted from $\mathbf{X}$ as a vector $\mathbf{\tilde{f}}_j \in \mathbb{R}^{n \times 1}$. These feature vectors are then combined to form the matrix $\mathbf{\hat{X}} \in \mathbb{R}^{n \times m}$, structured as follows:

\begin{equation}
\mathbf{\hat{X}} = [\mathbf{\tilde{f}}_1, \mathbf{\tilde{f}}_2, \dots, \mathbf{\tilde{f}}_{17}].
\label{eq:original_matrix}
\end{equation}

To perform PCA, we normalize $\mathbf{\hat{X}}$ using the corresponding feature-wise means $\bar{f}_j$ and standard deviations $\sigma_{f_j}$. The normalized matrix $\mathbf{\ddot{X}}$ is expressed as:

\begin{equation}
\mathbf{\ddot{X}} =
\begin{bmatrix}
\frac{\mathbf{\tilde{f}}_1 - \bar{f}_1}{\sigma_{f_1}},
\frac{\mathbf{\tilde{f}}_2 - \bar{f}_2}{\sigma_{f_2}},
\ldots,
\frac{\mathbf{\tilde{f}}_j - \bar{f}_j}{\sigma_{f_j}},
\ldots,
\frac{\mathbf{\tilde{f}}_{17} - \bar{f}_{17}}{\sigma_{f_{17}}}
\end{bmatrix}.
\label{eq:normalized_matrix}
\end{equation}

\noindent The PCA transformation is then applied to $\mathbf{\ddot{X}}$, yielding a set of principal components ordered by the amount of variance they explain.

We interpret the results of PCA by analyzing how the first two principal components are related to the meteorological features and their contributions to explaining the dataset's variability. Additionally, we evaluate the significance of the first two principal components for individual observations.

\subsubsection{Feature Contributions to Principal Components}
\begin{figure}[h]
    \centering
    \includegraphics[width=\textwidth]{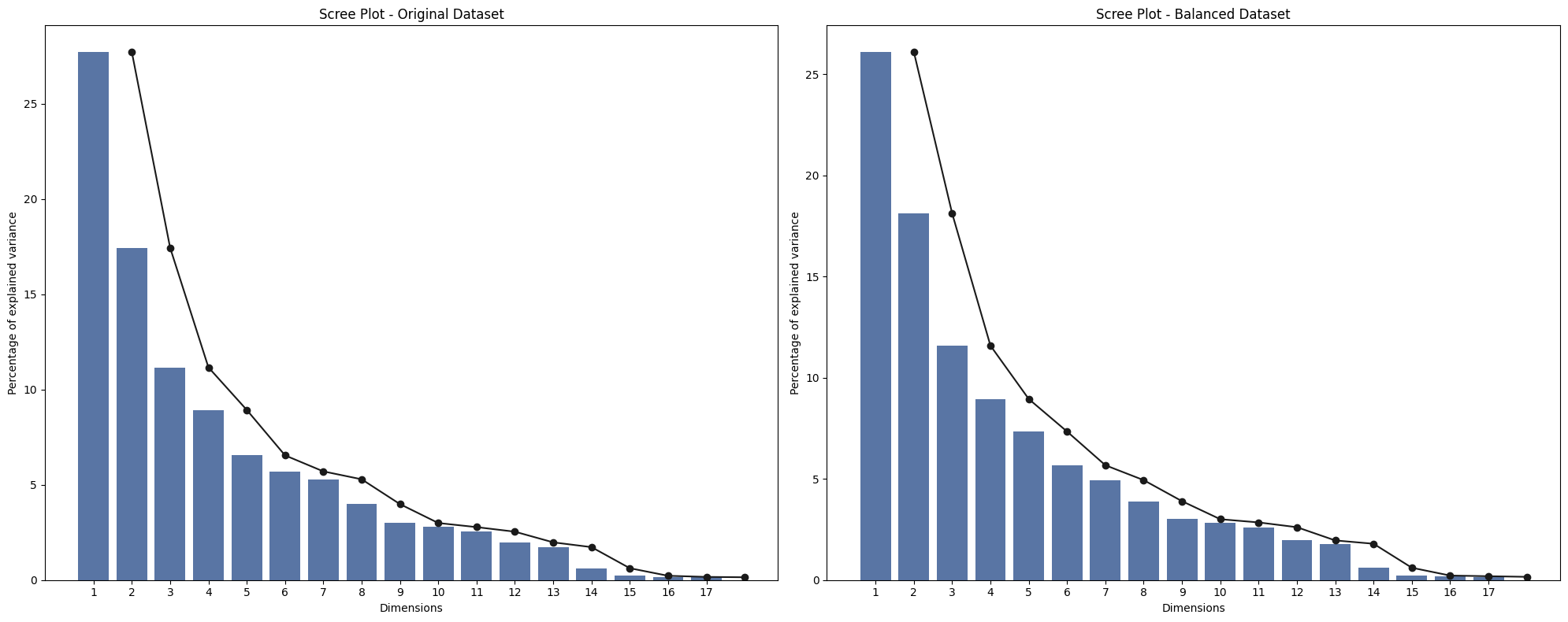}
    \caption{Percentage of variance explained by different principal components. We show the scree plot for original and balanced datasets.}
    \label{fig:biplots}
\end{figure}

To determine the components that account for the majority of variance in the dataset, we used screen plots in Figure \ref{fig:biplots}. It illustrates the proportion of variance explained by each principal component in the original dataset, revealing that the contribution of individual components to the overall variance is relatively small. Out of the 17 components, the first 13 components are necessary to explain 98.84\% of the variance in the original dataset. However, the original dataset is imbalanced, as it contains more samples with negative labels than positive ones. To balance the dataset, all positive samples are preserved, and an equal number of negative samples are randomly selected from the remaining data. This process ensures an even distribution of positive and negative labels. Since principal components are orthogonal to one another, they are uncorrelated, allowing each component to explain a distinct underlying phenomenon in the data. For the balanced dataset, the first 13 components account for 98.82\% of the variance as Figure \ref{fig:biplots} shows. As shown in Figure \ref{fig:biplots}, the variance distribution and scree plot illustrate how the principal components capture different aspects of the dataset, highlighting the impact of the original dataset's imbalance on its subspace representation. The contribution of each variable to the principal components can be evaluated by variable loadings, which indicate the significance of each variable in each component.

Table \ref{feature_contributions_combined} lists the contribution of each variable to the first two principal components, highlighting features such as MaxTemp, Temp3pm, and MinTemp with strong positive contributions to PC1, emphasizing the significance of temperature-related variables. Variables with loadings above this threshold are shown in bold. Similarly, wind-related features like WindGustSpeed and WindSpeed3pm show substantial contributions to PC2, emphasizing their role in the variability captured by this principal component. As highlighted earlier, these variables are important for rainfall prediction, suggesting that the first principal component, which is heavily influenced by these variables, may play a key role in predicting rainfall. The following sections will delve into the significance of these principal components in rainfall prediction.

\begin{table}[htbp]
\centering
\caption{Contributions of meteorological features to the first two principal components}
\label{feature_contributions_combined}
\resizebox{\textwidth}{!}{%
\begin{tabular}{lccccccccc}
\toprule
 & \textbf{Location} & \textbf{MinTemp} & \textbf{MaxTemp} & \textbf{Rainfall} & \textbf{WindGustDir} & \textbf{WindGustSpeed} & \textbf{WindDir9am} & \textbf{WindDir3pm} & \textbf{WindSpeed9am} \\
\midrule
\textbf{PC1 Contribution} & -0.0067 & 0.3639 & 0.4197 & -0.0227 & -0.0960 & 0.1469 & -0.0897 & -0.0822 & 0.1157 \\
\textbf{PC2 Contribution} & 0.0273 & 0.0649 & -0.1465 & 0.2513 & 0.2478 & 0.3897 & 0.2298 & 0.2386 & 0.2937 \\
\midrule
 &  & \textbf{WindSpeed3pm} & \textbf{Humidity9am} & \textbf{Humidity3pm} & \textbf{Pressure9am} & \textbf{Pressure3pm} & \textbf{Temp9am} & \textbf{Temp3pm} & \textbf{RainToday} \\
\midrule
\textbf{PC1 Contribution} &  & 0.1236 & -0.2806 & -0.2098 & -0.2524 & -0.2827 & 0.4177 & 0.4105 & -0.0852 \\
\textbf{PC2 Contribution} &  & 0.3348 & 0.1102 & 0.2327 & -0.3412 & -0.2742 & -0.0273 & -0.1708 & 0.3182 \\
\bottomrule
\end{tabular}%
}
\end{table}

\subsubsection{Relation between principal components with RainTommorrow attributes}
\begin{figure*}[htbp]
    \centering
    \begin{minipage}[b]{0.48\textwidth}
        \centering
        \includegraphics[width=\textwidth]{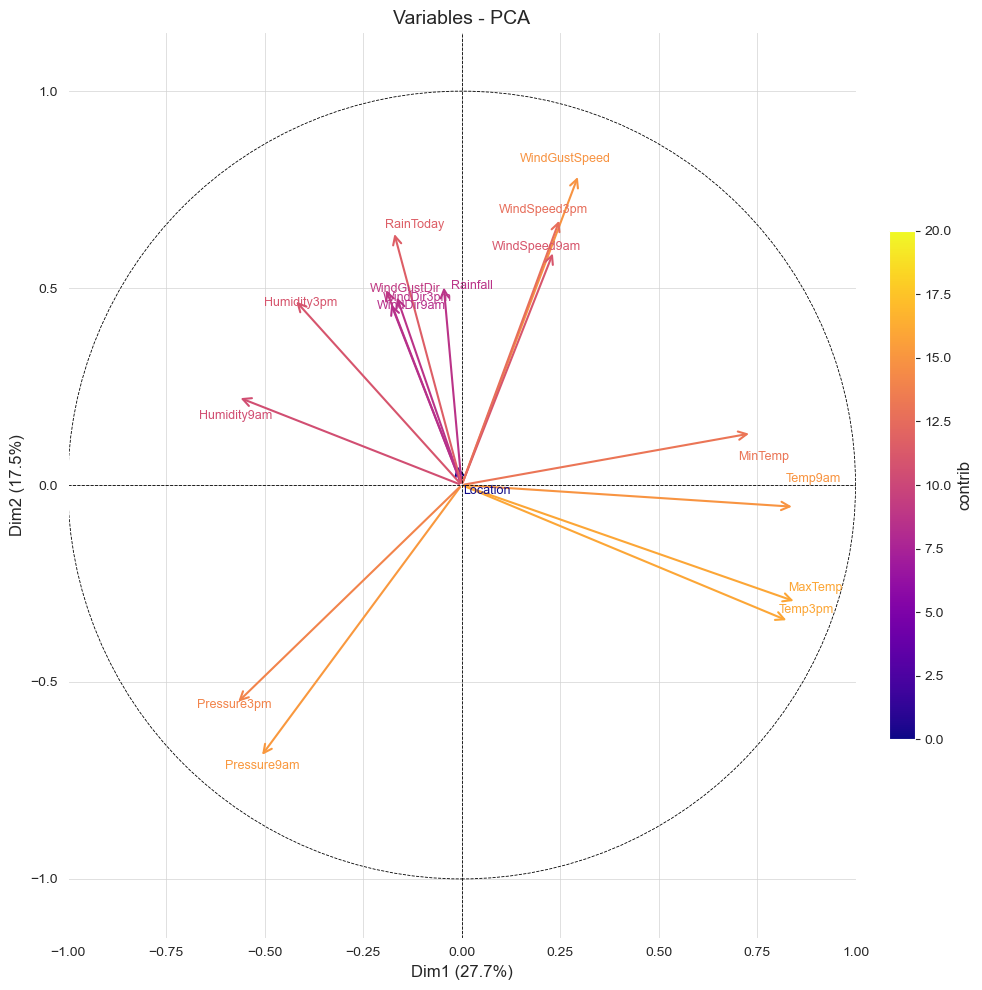}
        \label{fig:biplot_original}
    \end{minipage}
    \emph{   }\emph{   }\emph{   }
    \begin{minipage}[b]{0.48\textwidth}
        \centering
        \includegraphics[width=\textwidth]{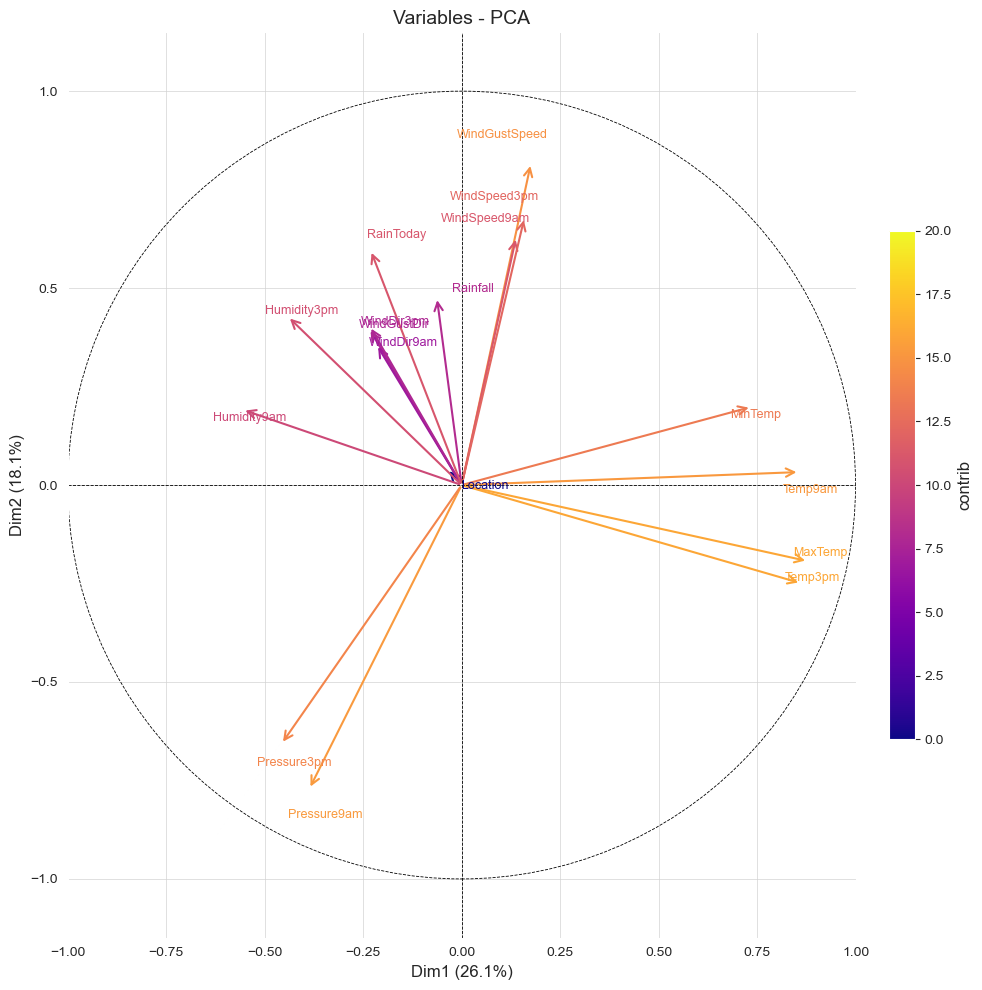}
        \label{fig:biplot_balanced}
    \end{minipage}
    \caption{Comparison of biplot representations for the original (left) and balanced (right) datasets. The biplots illustrate the contributions of meteorological features to the first two principal components PC1 \& PC2 in both datasets.}
    \label{fig:biplots_combined}
\end{figure*}

In this study, PCA was applied to analyze the relationships and contributions of meteorological features in both original and balanced datasets for predicting rainfall. The PCA biplots as shown in Figure \ref{fig:biplots_combined} visualize the feature contributions to the first two principal components, which explain a significant proportion of the variance in the dataset—26.1\% for PC1 and 18.1\% for PC2. The arrows in the biplot represent the original features, with their lengths indicating the magnitude of their contributions and their directions reflecting their relationships with the principal components. Temperature-related features, including MaxTemp, Temp3pm, and MinTemp, show strong positive contributions to PC1, indicating their central role in capturing the dataset's variability. Similarly, wind-related features, such as WindGustSpeed and WindSpeed3pm, dominate PC2, highlighting their distinct influence. Features near the origin, such as RainToday and Humidity3pm, have relatively minor contributions, while those closer to the circumference, such as temperature features, contribute significantly, as highlighted by the unit variance circle. The left biplot in Figure \ref{fig:biplots_combined} for the original dataset shows that temperature-related features are the most influential, as indicated by their long vectors extending towards the circumference. These features are highly correlated, which is consistent with their similar contributions to the principal components. In contrast, wind speed and precipitation features exhibit shorter vectors, suggesting their limited influence on the explained variance. Furthermore, temperature and humidity demonstrate a positive correlation, while wind speed and precipitation appear orthogonal, suggesting that they capture complementary information. For the balanced dataset, the right biplot in Figure \ref{fig:biplots_combined} shows the contributions of features such as humidity and wind speed become more prominent. The balancing process redistributes data points more evenly along the principal components, enhancing feature representation and improving class separation. Despite these adjustments, the overall orientation of feature vectors remains consistent with the unbalanced dataset, confirming the robustness of the underlying feature relationships.

These findings emphasize the role of temperature-related features, such as MaxTemp, Temp3pm, and MinTemp, in defining the dataset's variability. Additionally, balancing the dataset enhances the representation of underrepresented features, such as humidity, wind speed, and precipitation. This result also validates the effectiveness of constructing new features, such as maximum temperature difference and maximum humidity, as they contribute significantly to the principal components.
\subsubsection{Relation of principal components with individual data}
\begin{figure*}[htbp]
    \centering
    \includegraphics[width=0.7\textwidth]{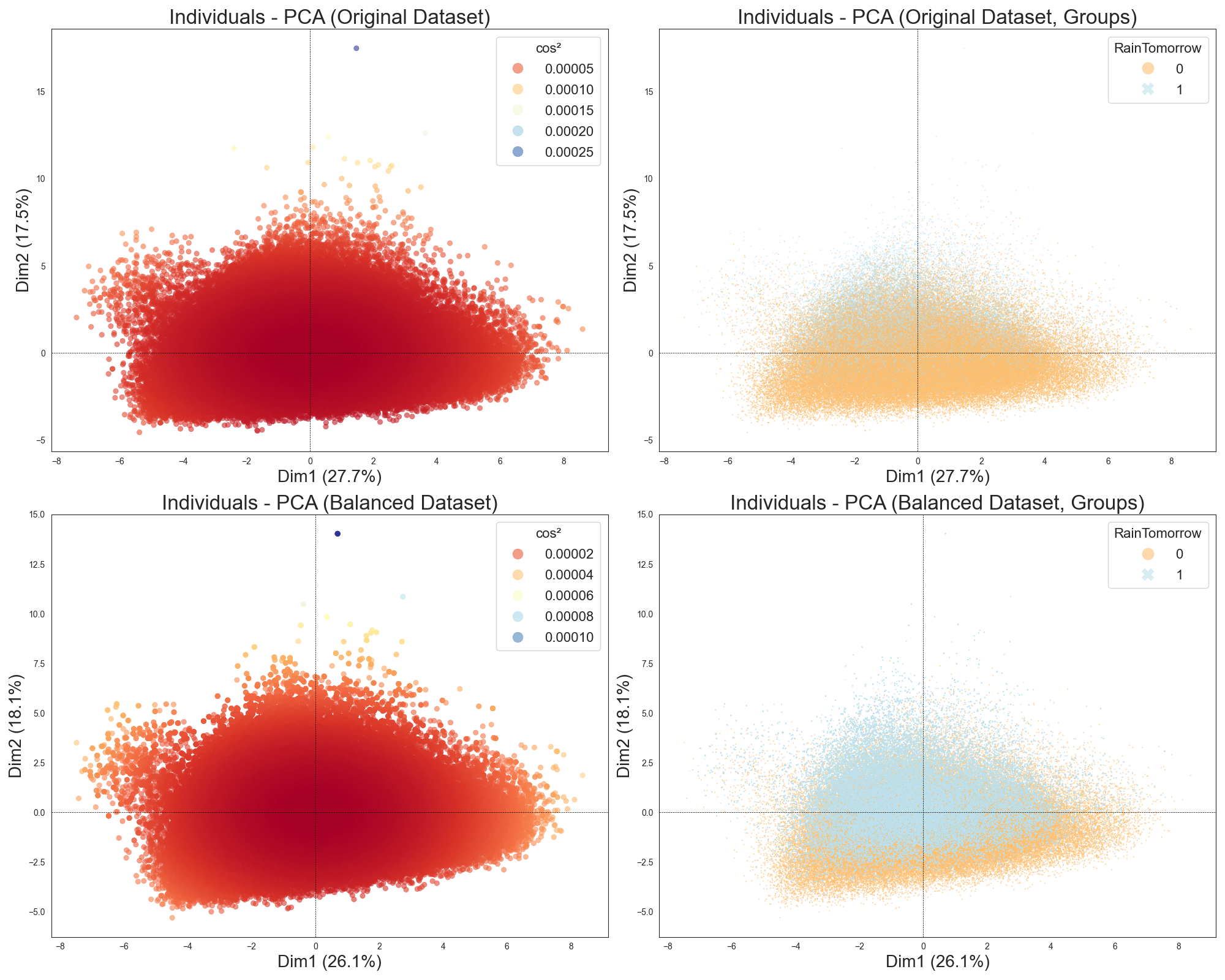}
    \caption{PCA subspace representation of observations for the first two principal components. For the original and balanced datasets, color-coded by cos2 values are shown in the left column and RainTomorrow label groups are demonstrated in the right column.}
    \label{fig:scatter_plots}
\end{figure*}

To further investigate the PCA-transformed data, scatter plots of individual observations are shown in Figure \ref{fig:scatter_plots}, comparing the original and balanced datasets. The upper-left panel represents the original dataset, with point colors indicating cos2 values that measure the quality of representation in the PCA space. Points with higher cos2 values, depicted in red hues, are better captured by the first two principal components, while lower cos2 values suggest incomplete representation. In contrast, the lower-left panel shows the balanced dataset with cos2-based coloring. After balancing, the points are more evenly distributed, demonstrating enhanced representation of underrepresented classes.
The upper-right and lower-right panels illustrate the distribution of observations by the RainTomorrow variable. In the original dataset, rainy and non-rainy days show considerable overlap, indicating poor class separation. However, the balanced dataset shows improved separation along the principal components, particularly for Dim1, which explains 27.7\% of variance in the original dataset and 26.1\% in the balanced dataset. Dim2 contributes additional discriminative power, enabling clearer differentiation between classes. The class balancing improves the feature space and enhances the separation of rainy and non-rainy days in the PCA-transformed data.

Overall in this section, we applied data preprocessing and feature engineering to address missing values, outliers, and imbalances in meteorological data. Constructed features, such as maximum temperature difference and maximum humidity, proved effective in enhancing the dataset's variability representation. PCA and t-SNE methods facilitated dimensionality reduction and cluster analysis, validating the relevance of selected features. PCA highlighted the critical role of temperature and wind-related features in explaining dataset variability, while class balancing improved the representation of underrepresented features and enhanced class separability. In the next section, we outline the methodology for training and evaluating various machine learning and deep learning models for rainfall prediction. We detail the experimental setup, including data partitioning, hyperparameter tuning, and model configurations. Finally, we compare model performances using metrics like accuracy and ROC-AUC to identify the best-performing approach.

\newpage
\section{Model} 
\label{sec:training}
\begin{figure}[H]
    \centering
    \begin{minipage}[b]{0.24\textwidth}
        \includegraphics[width=\textwidth]{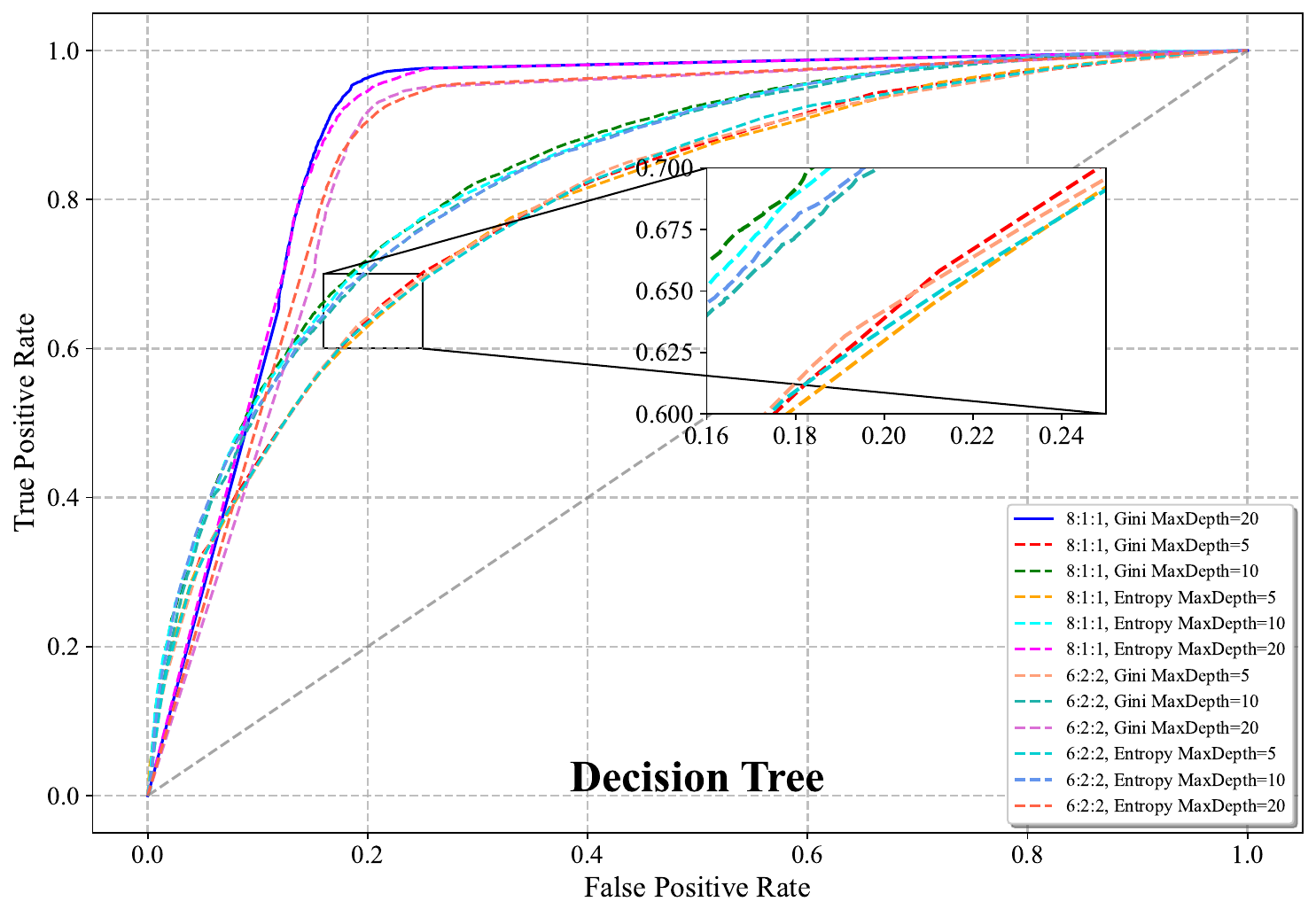}
    \end{minipage}
    \begin{minipage}[b]{0.24\textwidth}
        \includegraphics[width=\textwidth]{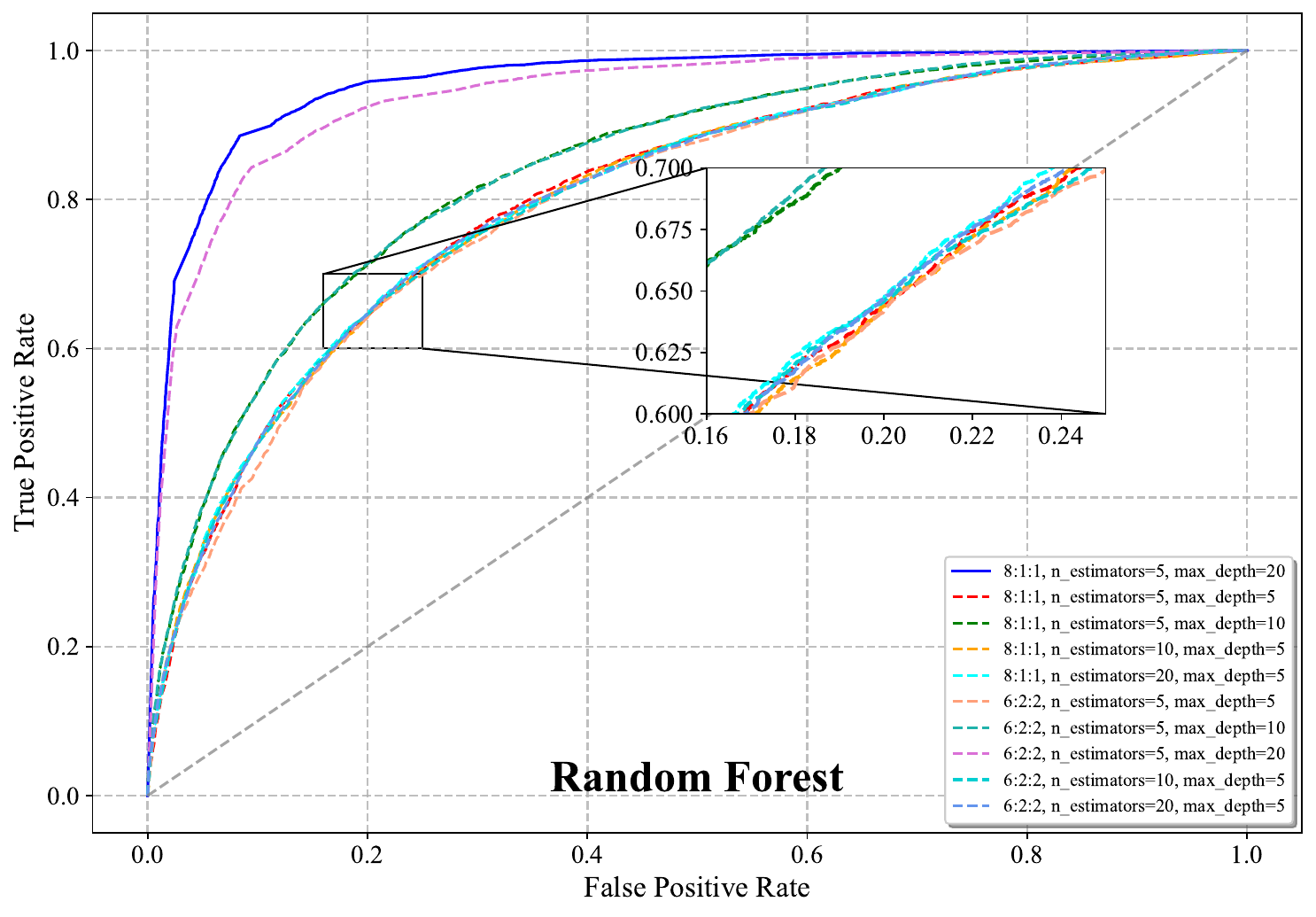}
    \end{minipage}
    \begin{minipage}[b]{0.24\textwidth}
        \includegraphics[width=\textwidth]{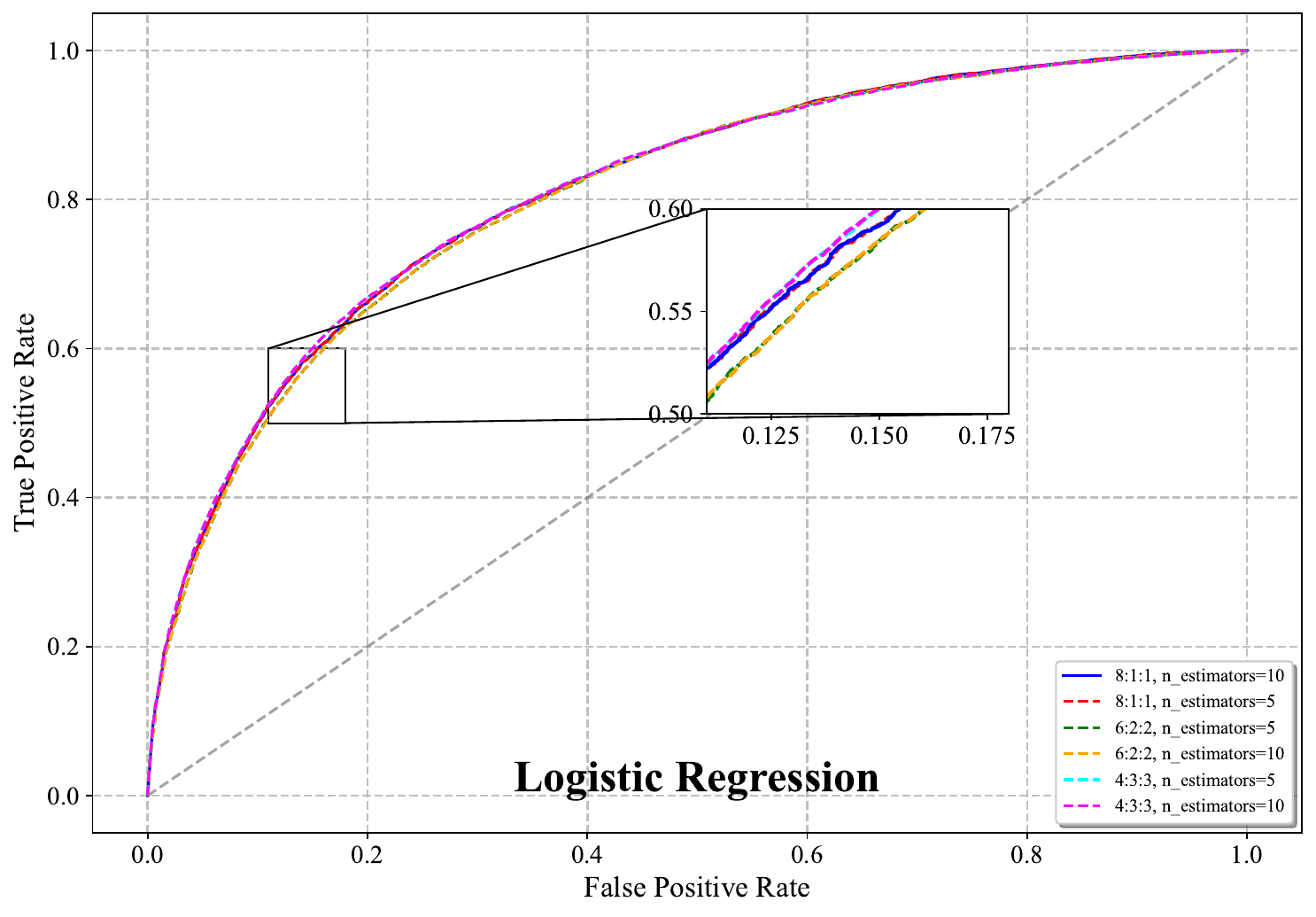}
    \end{minipage}
    \begin{minipage}[b]{0.24\textwidth}
        \includegraphics[width=\textwidth]{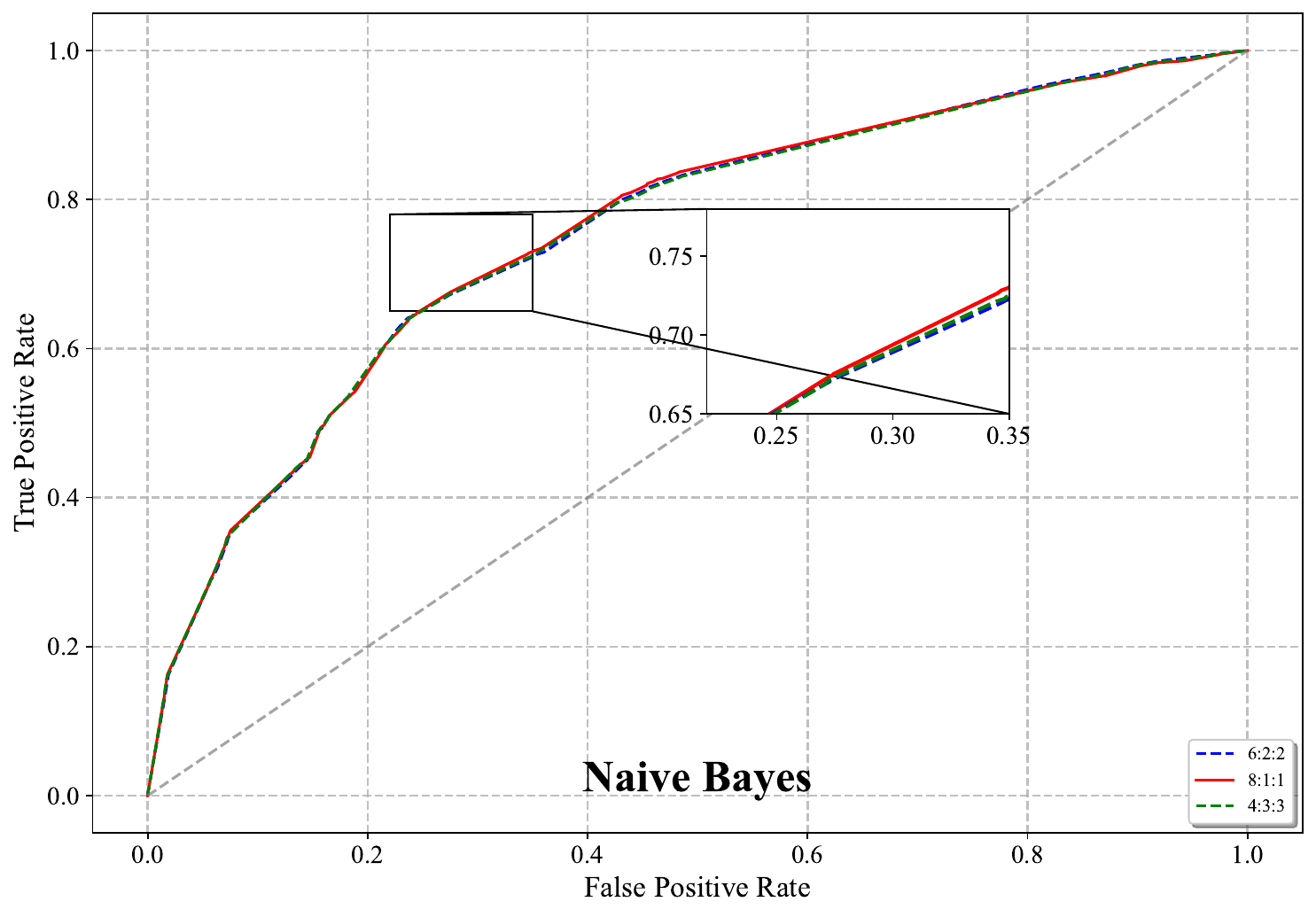}
    \end{minipage}

    \vspace{0.00cm} 
    \begin{minipage}[b]{0.24\textwidth}
        \includegraphics[width=\textwidth]{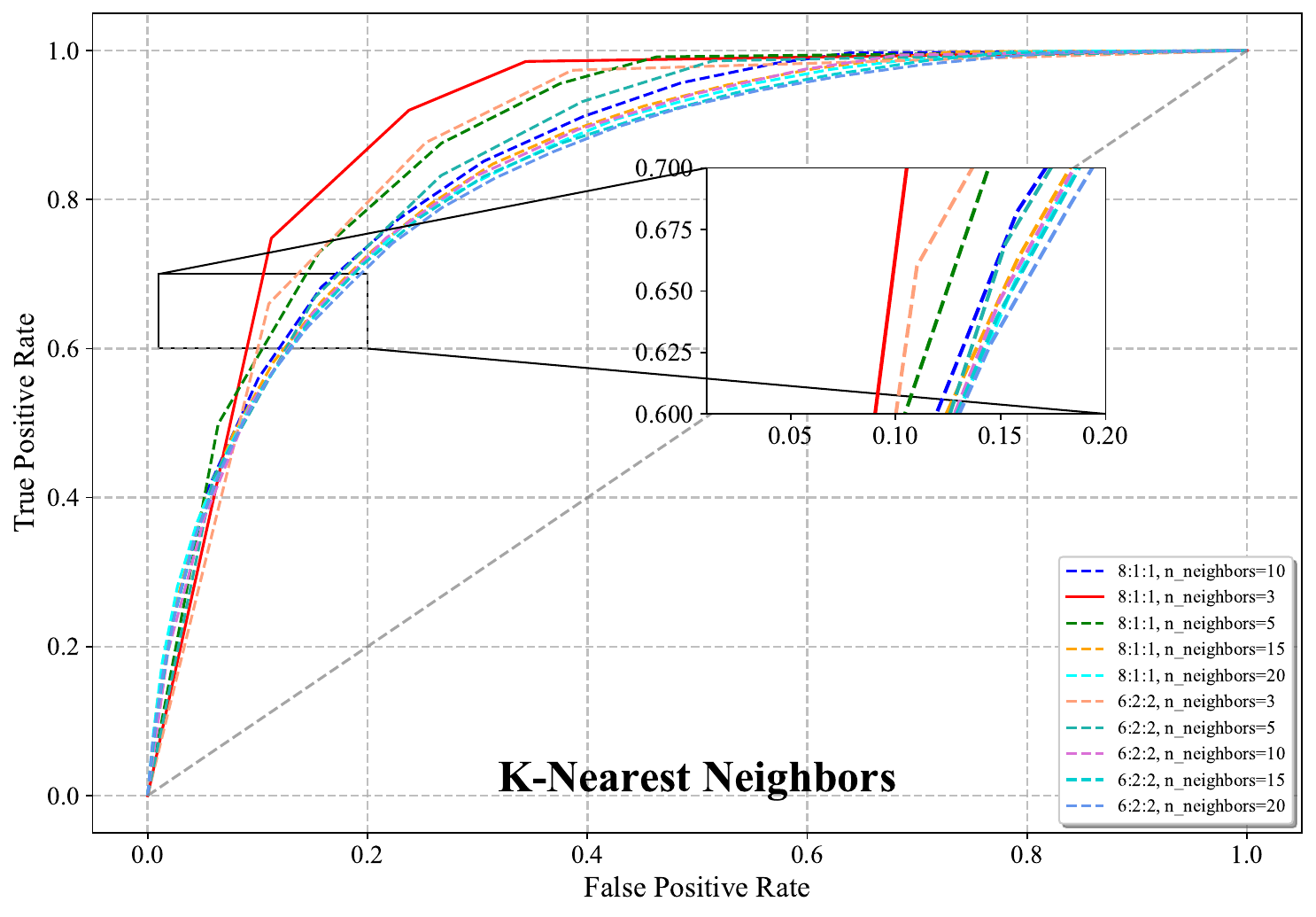}
    \end{minipage}
    \begin{minipage}[b]{0.24\textwidth}
        \includegraphics[width=\textwidth]{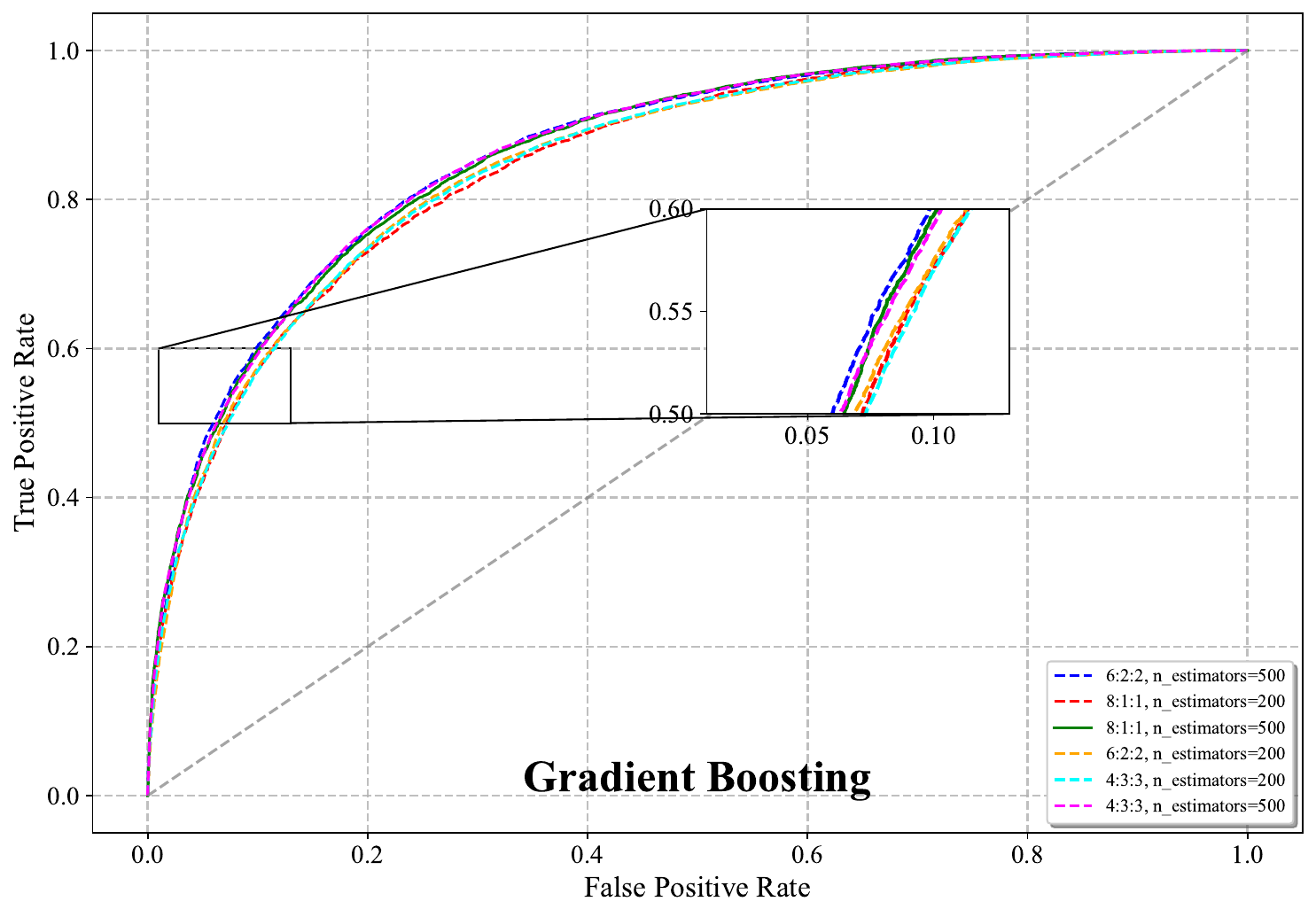}
    \end{minipage}
    \begin{minipage}[b]{0.24\textwidth}
        \includegraphics[width=\textwidth]{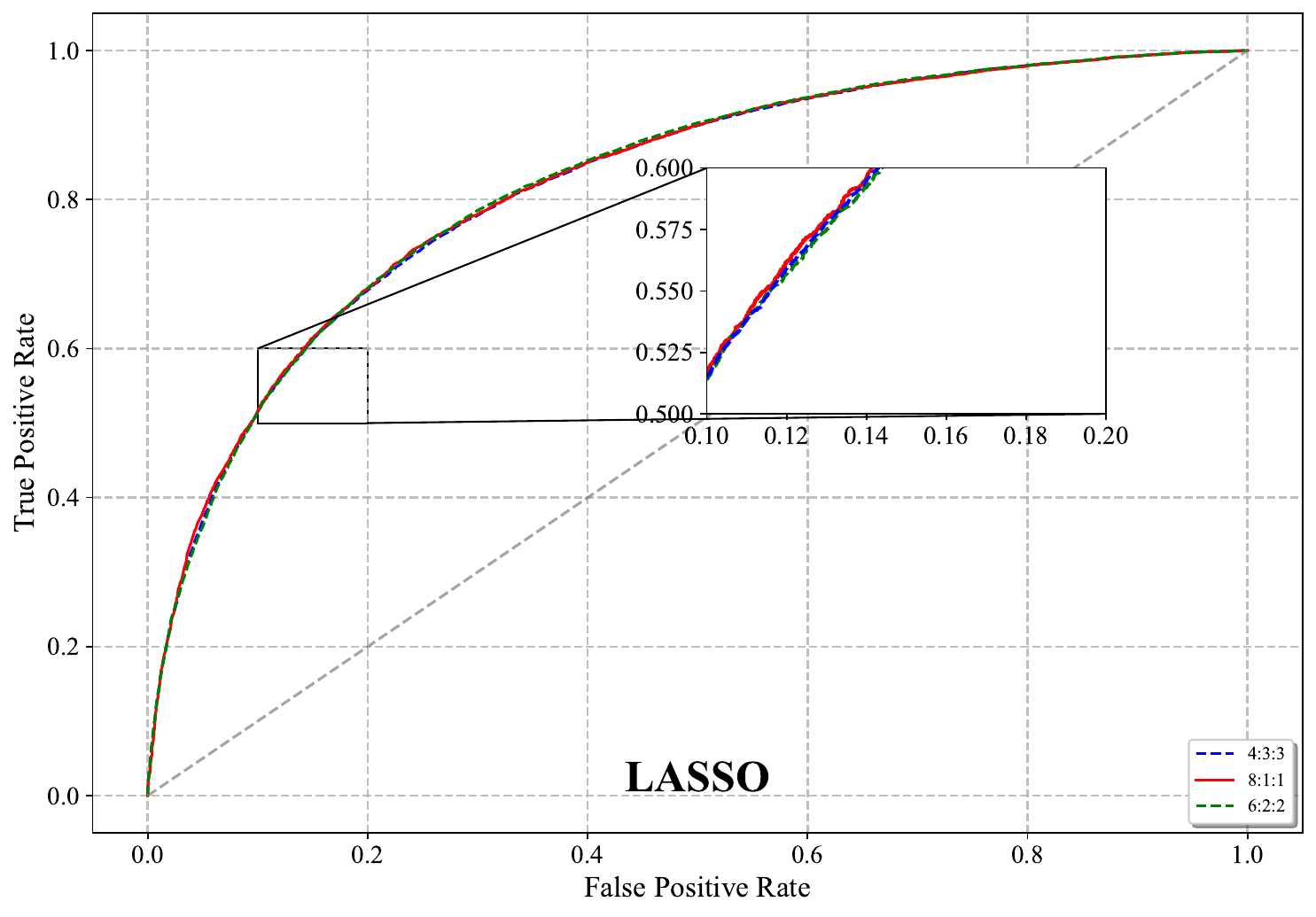}
    \end{minipage}
    \begin{minipage}[b]{0.24\textwidth}
        \includegraphics[width=\textwidth]{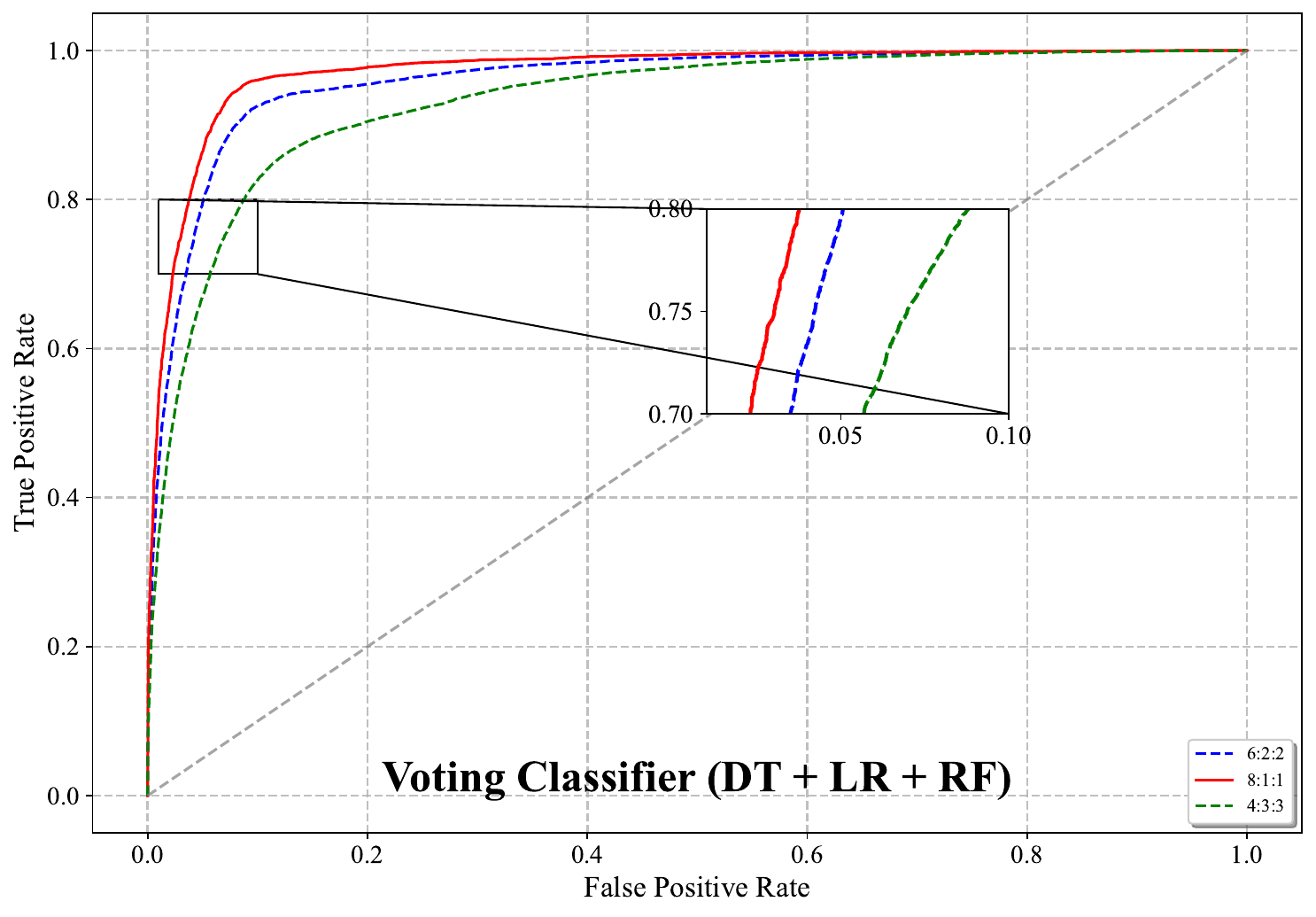}
    \end{minipage}

    \vspace{0.00cm} 
    \begin{minipage}[b]{0.24\textwidth}
        \includegraphics[width=\textwidth]{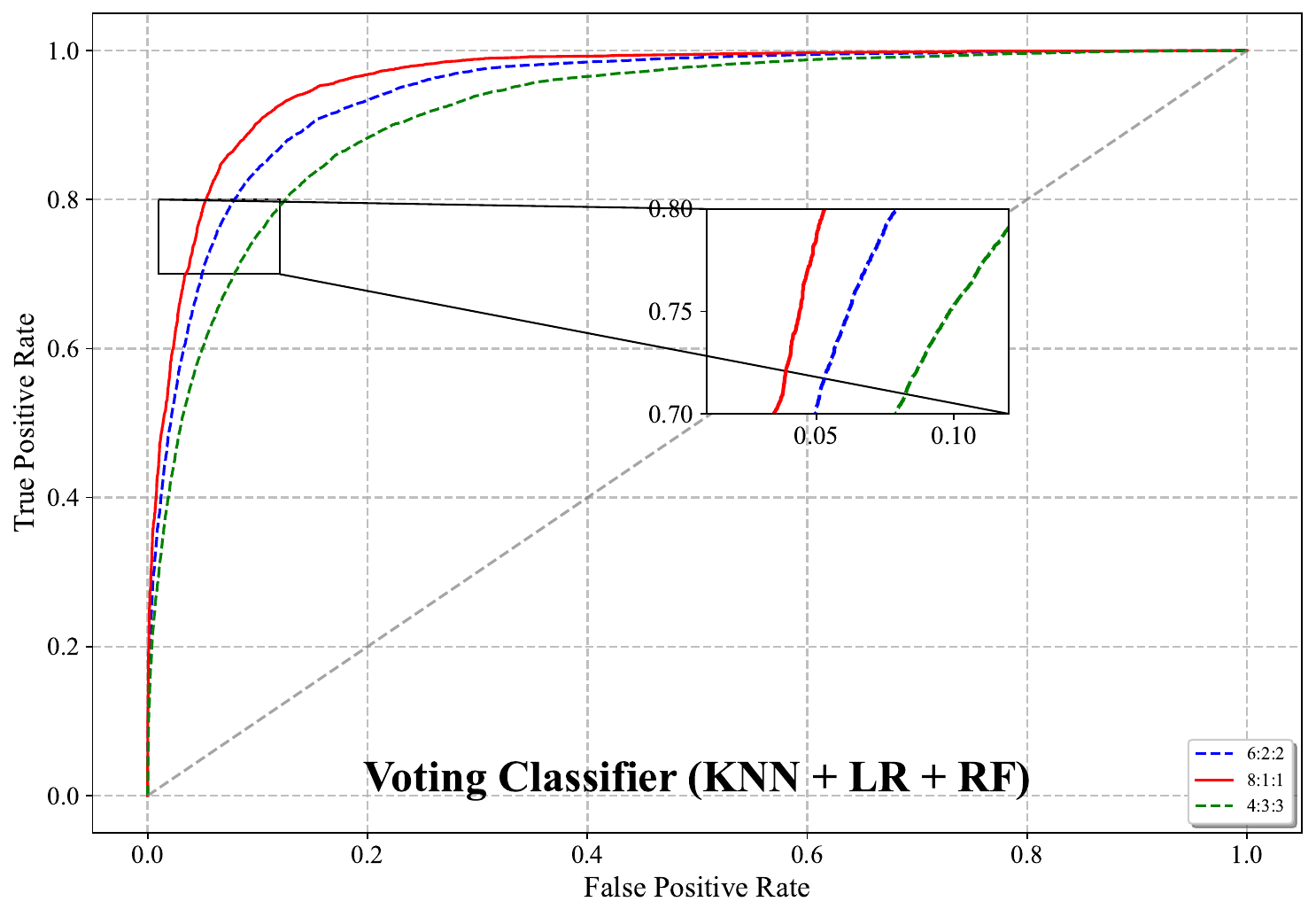}
    \end{minipage}
    \begin{minipage}[b]{0.24\textwidth}
        \includegraphics[width=\textwidth]{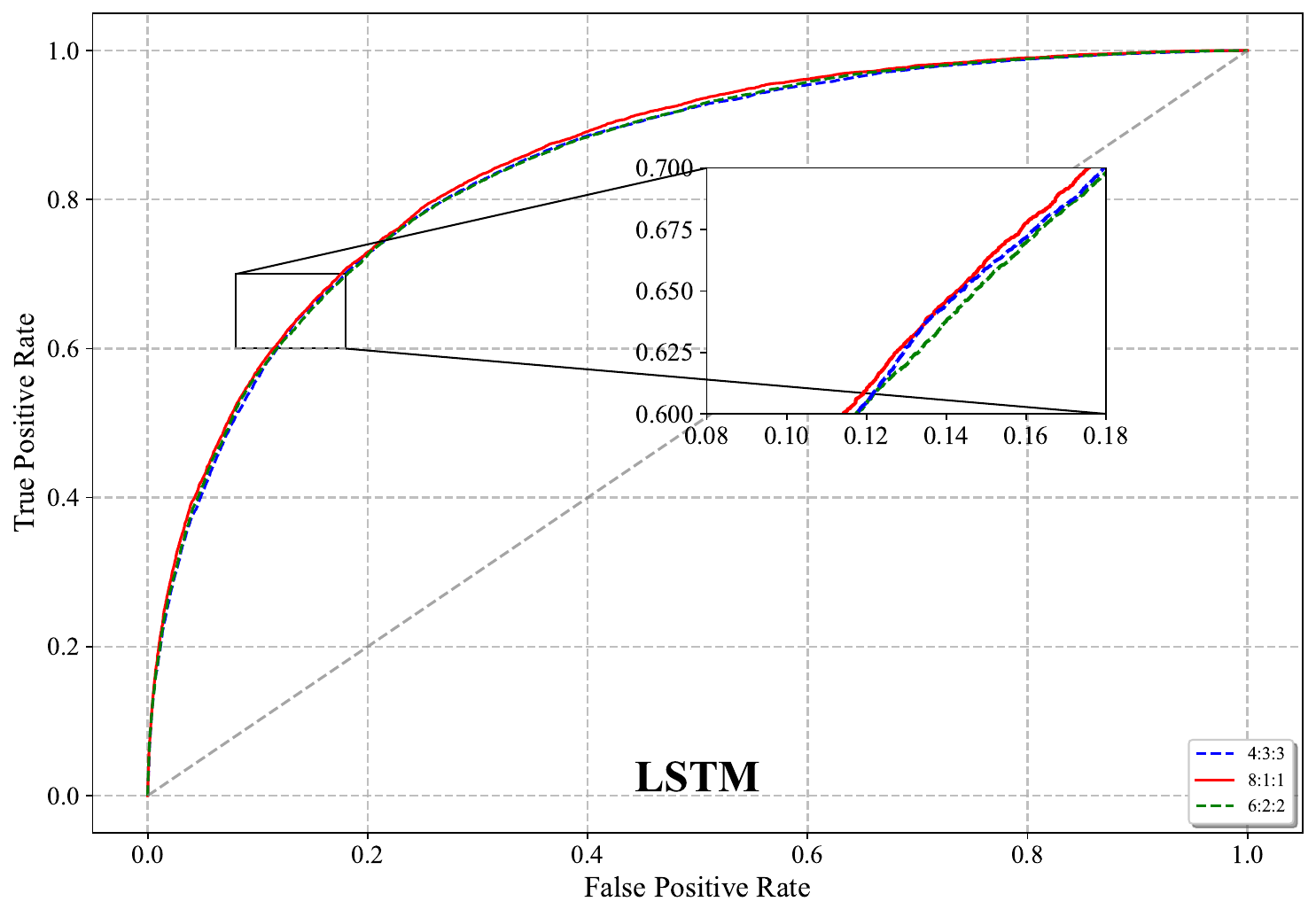}
    \end{minipage}
    \begin{minipage}[b]{0.24\textwidth}
        \includegraphics[width=\textwidth]{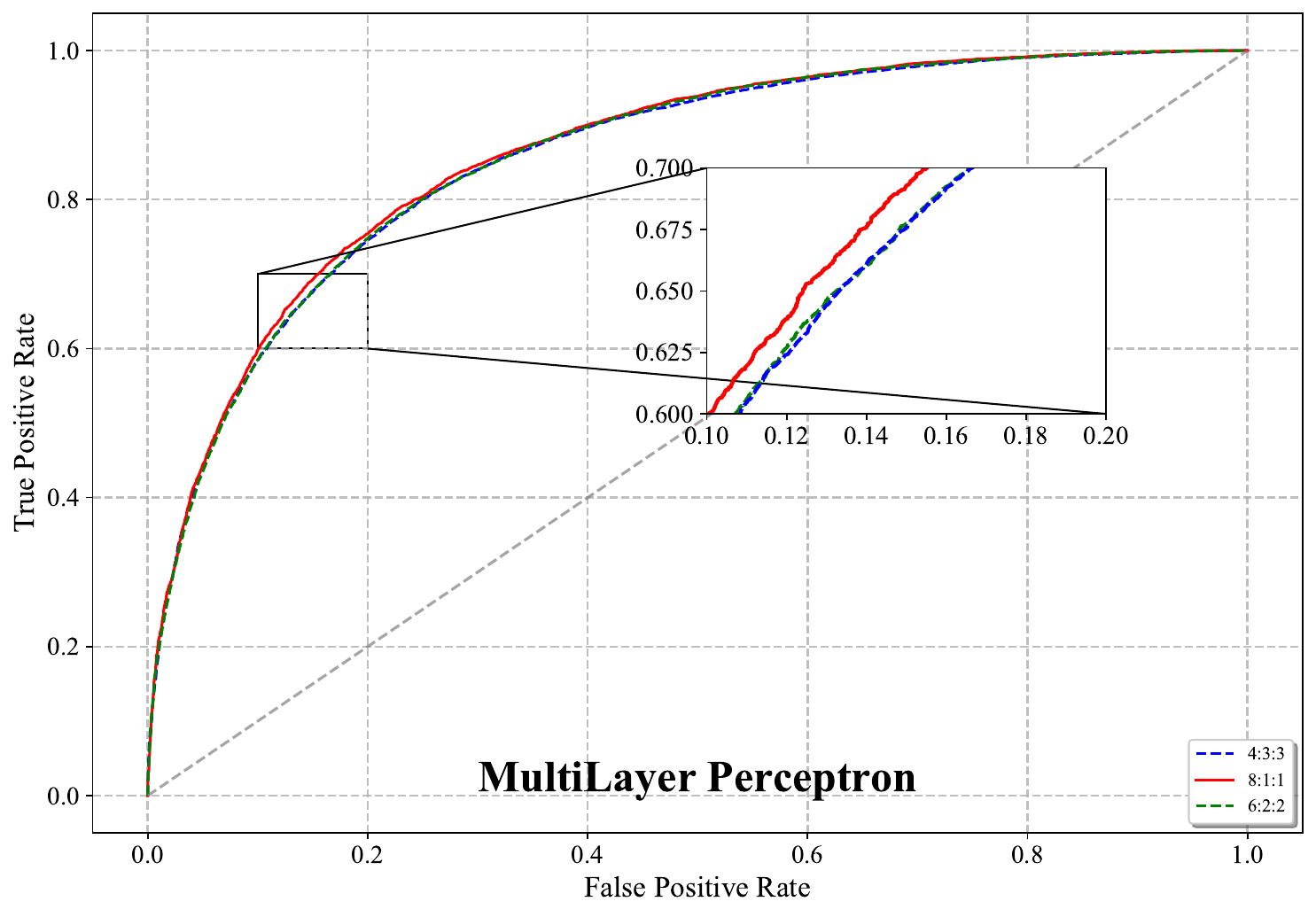}
    \end{minipage}
    \begin{minipage}[b]{0.24\textwidth}
        \includegraphics[width=\textwidth]{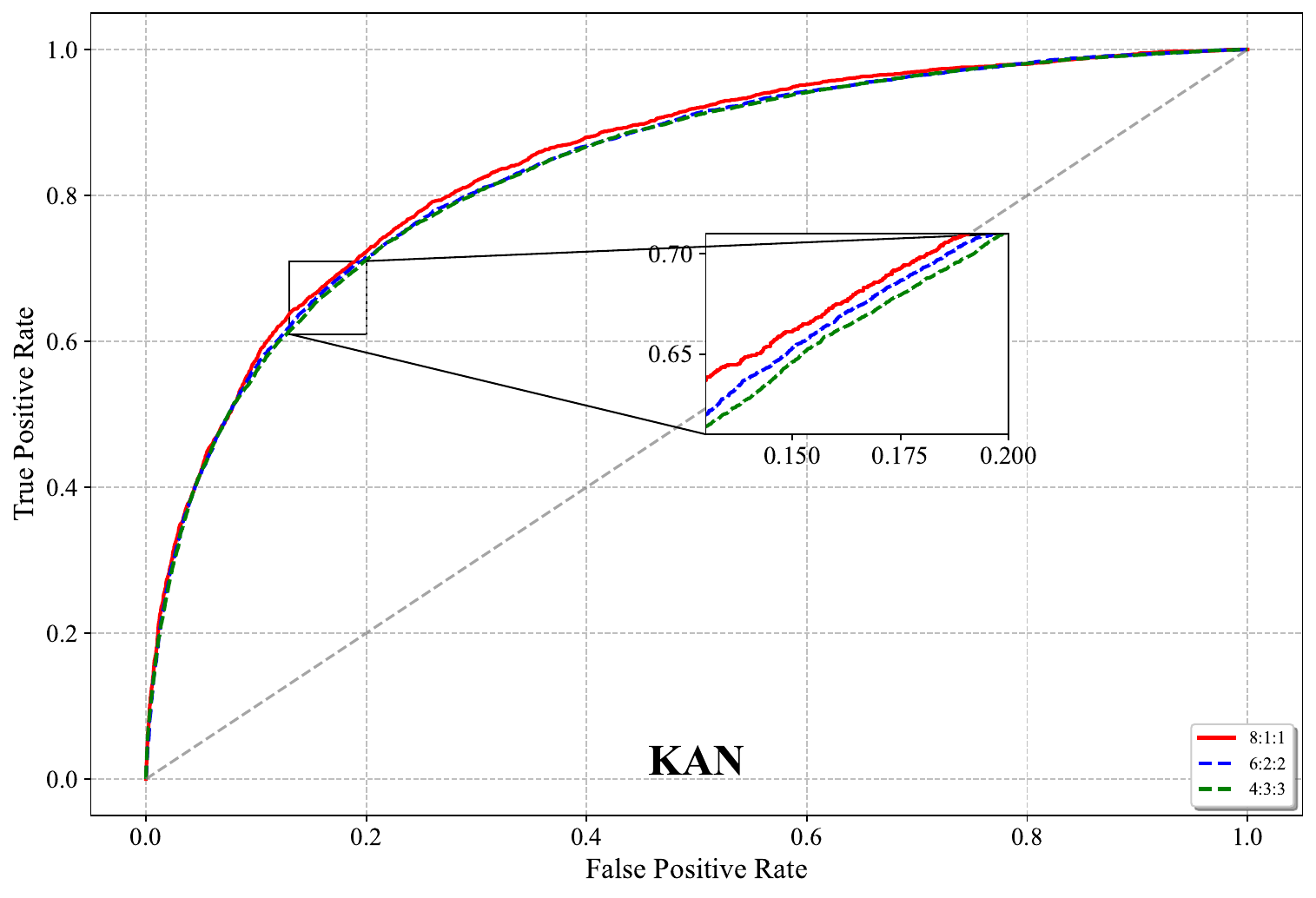}
    \end{minipage}
    \caption{ROC curves for weak/advanced ML models under different specific model settings and training settings (e.g., 8:1:1, 6:2:2, and 4:3:3 dataset split ratios), aimed at identifying patterns in model settings to provide reliable ranges for subsequent grid search and selecting the optimal dataset split ratios. Each sub-figure represents a specific model such as Decision Tree, Random Forest, Logistic Regression, Naive Bayes, KNN, Gradient Boosting, LASSO, LSTM, MLP, KAN, etc.}
    \label{fig:preliminary_experimentsroc}
\end{figure}

Our proposed approach RAINER includes non-learning mathematical-based approaches such as Linear Discriminant Analysis (LDA) and linear regression; Learning-based methods contain both weak machine learning classifiers such as Decision Tree (DT), Random Forest (RF), Logistic Regression (LR), Naive Bayes (NB), K-Nearest Neighbors (KNN), Gradient Boosting (GB), LASSO, ElasticNet, and ensemble models like Voting Classifiers (e.g., DT+LR+RF and KNN+LR+RF), as well as advanced learning-based models, including Multilayer Perceptron (MLP), CNN, Graph Convolutional Networks (GCN), ResNet, RNN, LSTM, Transformers, KAN, and voting combinations such as MLP+LSTM+Transformer, KAN+LSTM+Transformer, and KAN+ResNet. We first conducted a \textbf{pre-exploration experiment}, as shown in Figure \ref{fig:preliminary_experimentsroc},  to analyze the impact of hyperparameters on model performance and identify consistent trends. These trends enabled us to determine more precise and reliable parameter ranges, effectively guiding subsequent grid search and optimization processes. Additionally, this pre-exploration allowed us to standardize the training settings, such as dataset split ratios. Based on our experiments, we observed that the 8:1:1 dataset split ratio consistently outperformed others (e.g., 6:2:2 and 4:3:3). Therefore, we adopted the 8:1:1 split ratio as the standard configuration for all subsequent experiments.

Based on the identified patterns, we provided refined parameter ranges for grid search, increasing the likelihood of identifying optimal settings while avoiding sub-optimal solutions caused by randomly defined ranges. By focusing on well-informed parameter ranges, we minimized the risk of the algorithm getting stuck in a local maximum. We applied GridSearchCV for each model to systematically explore parameter combinations. This approach, combining grid search and cross-validation, ensured that each parameter was evaluated iteratively within the specified range, and the best combination was selected based on its performance on the validation set. Although grid search is exhaustive and computationally expensive—particularly with large datasets and numerous parameters—our pre-exploration step significantly reduced the search space, making the process more efficient and effective. Table \ref{Optimal hyperparameter settings for each model derived via grid search optimization} shows the result of optimal settings for each model after grid search. Then we will utilize the optimal model and training setting to conduct several quantitative experiments across four different feature sets and three categories of methods in the next section.

\begin{table}[h!]
\centering
\resizebox{\textwidth}{!}{
\begin{tabular}{|l|l|}
\hline
\multicolumn{2}{|c|}{\textbf{Non-learning Mathematical Approaches}} \\
\hline
LDA &solver = `svd', shrinkage = None, n\_components = None, store\_covariance = False, tol = 0.0001 \\
LinearRegression & fit\_intercept = True, normalize = False, copy\_X = True, n\_jobs = None \\
\hline
\multicolumn{2}{|c|}{\textbf{Weak Machine Learning Classifier}} \\
\hline
LASSO & C = 500, penalty = `l\_1', solver = `liblinear'\\
ElasticNet & C = 100, l1\_ratio = 0.001\\
DT & criterion = `gini', max\_depth = 40, max\_features = `sqrt', min\_samples\_leaf = 1, min\_samples\_split = 2\\
NB & alpha = 0.1, binarize = 0\\
KNN & n\_neighbors = 100, p = 1, weight = `distance'\\
RF & n\_estimators = 100, max\_depth = None\\
LR  & estimator\_C = 10, estimator\_max\_iter = 1000, n\_estimators = 50\\
DT+LR+RF  & (criterion = `gini', max\_depth = 40, max\_features = `sqrt', min\_samples\_leaf = 1, min\_samples\_split = 2), (estimator\_C = 10, estimator\_max\_iter = 1000, n\_estimators = 50), (n\_estimators = 100, max\_depth = None)\\
KNN+LR+RF  & (n\_neighbors = 100, p = 1, weight = `distance'),  (estimator\_C = 10, estimator\_max\_iter = 1000, n\_estimators = 50), (n\_estimators = 100, max\_depth = None)\\
\hline
\multicolumn{2}{|c|}{\textbf{Advanced Learning-based Model}} \\
\hline
MLP  & learning\_rate = 0.001, hidden\_sizes = [128, 64, 32], activation = `tanh', alpha = 0.001,\\
CNN  & learning\_rate = 0.001, batch\_size = 64, num\_filters = 32, kernel\_size = 3, activation = `relu' \\
GNN  &learning\_rate = 0.001, batch\_size = 64, hidden\_dim = 16, activation = `relu'  \\
ResNet  & learning\_rate = 0.001, batch\_size = 64, num\_blocks = 2, hidden\_dim = 32, kernel\_size = 3, activation = `relu' \\
RNN  & learning\_rate = 0.001, hidden\_size = 64, num\_layers = 2, activation = `relu'  \\
LSTM  & learning\_rate = 0.0005, batch\_size = 64, dropout\_rate = 0.2, hidden\_size = 50,  num\_layers = 1\\
Transformer  & learning\_rate = 0.001, embed\_dim = 16, nhead = 4, num\_layers = 1, dim\_feedforward = 64, activation = `relu'  \\
KAN  & learning\_rate = 0.001, Q = 2, hidden\_dim = 64, activation = `relu'  \\
MLP+LSTM+Transformer & (learning\_rate = 0.001, hidden\_sizes = [128, 64, 32], activation = `tanh'), (learning\_rate = 0.0005, batch\_size = 64, dropout\_rate = 0.2, hidden\_size = 50), (learning\_rate = 0.001, embed\_dim = 16, nhead = 4, , dim\_feedforward = 64, activation = `relu') \\
KAN+LSTM+Transformer & (learning\_rate = 0.001, Q = 2, hidden\_dim = 64, activation = `relu'), (learning\_rate = 0.0005, batch\_size = 64, dropout\_rate = 0.2, hidden\_size = 50), (learning\_rate = 0.001, embed\_dim = 16, nhead = 4, dim\_feedforward = 64, activation = `relu') \\
KAN+ResNet  & (learning\_rate = 0.001, Q = 2, hidden\_dim = 64, activation = `relu'), (learning\_rate = 0.001, batch\_size = 64, num\_blocks = 2, hidden\_dim = 32, kernel\_size = 3, activation = `relu') \\
\hline
\end{tabular}
}
\caption{Optimal hyperparameter settings for each model derived via grid search optimization}
\label{Optimal hyperparameter settings for each model derived via grid search optimization}
\end{table}

\section{Experiment} 
\subsection{Evaluation Metrics}

The evaluation metrics used to assess the model performance include Accuracy, Precision, Recall, Area Under the ROC Curve (AUC), and F1-score. These metrics provide a comprehensive analysis of both the overall and class-specific predictive capabilities. 

\begin{itemize}
\item \textbf{Accuracy}: it measures the overall correctness of the model predictions and is defined as:
\begin{equation}
\text{Accuracy} = \frac{\text{TP} + \text{TN}}{\text{TP} + \text{TN} + \text{FP} + \text{FN}},
\end{equation}
where TP, TN, FP, and FN represent the counts of true positives, true negatives, false positives, and false negatives, respectively. 

\item \textbf{Precision}: it is also referred to as the positive predictive value, quantifies the proportion of correctly predicted positive samples among all samples predicted as positive. It emphasizes the reliability of positive predictions and is particularly important in scenarios where false positives are costly. The formula is defined as below:
\begin{equation}
\text{Precision} = \frac{\text{TP}}{\text{TP} + \text{FP}}.
\end{equation}

\item \textbf{Recall}: it also represents sensitivity, and measures the model’s ability to identify all relevant positive samples. It captures the completeness of the positive predictions, which is critical in applications where missed positives carry significant consequences:

\begin{equation}
\text{Recall} = \frac{\text{TP}}{\text{TP} + \text{FN}}.
\end{equation}

\item \textbf{AUC}: To evaluate the trade-off between the true positive rate (TPR) and the false positive rate (FPR) across different decision thresholds, the AUC is employed. The AUC measures the overall ability of the model to discriminate between positive and negative classes. Mathematically, the AUC is defined as the area under ROC curve, and it can be expressed as:
\begin{equation}
\text{AUC} = \int_0^1 \text{TPR}(t) \, d\text{FPR}(t),
\end{equation}
where \( \text{TPR}(t) \) and \( \text{FPR}(t) \) represent the true positive rate and false positive rate at a given threshold \( t \), respectively. Alternatively, the AUC can also be interpreted as the probability that a randomly chosen positive sample (\( y^+ \)) has a higher predicted score (\( \hat{y}^+ \)) than a randomly chosen negative sample (\( y^- \)), which is given by:
\begin{equation}
\text{AUC} = P(\hat{y}^+ > \hat{y}^-).
\end{equation}

The ROC curve is constructed by plotting TPR against FPR at various thresholds, providing a visual representation of the model's classification performance across different operating points. A higher AUC value indicates better discrimination between the positive and negative classes, with an AUC of 1 representing a perfect classifier and an AUC of 0.5 indicating random guessing.

\item \textbf{F-measure}: it is often referred to as the \( F_{\beta} \)-score, is a weighted harmonic mean of precision and recall, where the parameter \( \beta \) determines the weight of recall relative to precision. When \( \beta = 1 \), the formula focuses more on recall while balancing it with precision. The general form of the \( F_{\beta} \)-score is expressed as:
\begin{equation}
F_{\beta} = (1 + \beta^2) \cdot \frac{\text{Precision} \cdot \text{Recall}}{\beta^2 \cdot \text{Precision} + \text{Recall}}.
\end{equation}
where we choose \( \beta = 1 \). The \( F_1 \)-score provides a balanced measure of model performance, making it particularly suitable for scenarios where both precision and recall are equally important.
\end{itemize}

\subsection{Quantitative Experiment}
The quantitative experiment shown in Table~\ref{Quantitative experiment on model performances under different feature engineering strategies} investigates the performance of various models—ranging from non-learning mathematical approaches, weak machine learning classifiers, to advanced learning-based models—under different feature engineering strategies. First, it is observed that weak machine learning classifiers, such as DT, RF, and KNN, outperform several advanced methods, particularly in optimized grid search settings. For example, RF achieves an accuracy of 85.2\% and AUC of 87.5\% with the original feature set, surpassing deep models like CNN and GCN. This highlights the robustness of weak classifiers in small to medium-scale problems where advanced models may face challenges such as overfitting or underfitting due to insufficient data complexity. The structured nature of rainfall prediction, involving well-defined meteorological features, makes these traditional classifiers particularly effective. Second, ensemble voting methods (DT+LR+RF and KNN+LR+RF) demonstrate significant performance gains across multiple metrics, achieving stability and balance. While the top-performing individual models often dominate in certain metrics (marked in red), ensemble methods consistently secure high results in others (marked in green and blue). For instance, DT+LR+RF achieves a precision of 87.1\% and recall of 97.3\% under the ``Selected + Constructed Features" strategy, showcasing the power of combining models to trade off individual strengths and weaknesses. Third, RF emerges as a standout model across all feature sets, consistently achieving top-tier results. This suggests that RF's ability to handle feature importance and interactions may contribute to its superior performance in rainfall prediction tasks. We hypothesize that RF efficiently leverages newly constructed features, such as maximum temperature and humidity differences, which capture critical weather dynamics. These features, along with PCA-reduced components, likely provide RF with a richer feature space for learning complex patterns.

For the comparison among different dataset feature strategies, we observe that our proposed "Selected + Constructed Features" strategy achieves the highest performance in most cases, even outperforming add-on PCA-based approaches. This indicates that manually constructed features retain critical information that PCA might overlook during dimensionality reduction. While PCA effectively reduces feature space and mitigates overfitting risks, it may inadvertently discard important feature relationships that contribute to model accuracy. The manually constructed features, on the other hand, preserve domain-specific insights, such as humidity and temperature variations, that are highly correlated with rainfall prediction tasks. This highlights the value of domain knowledge-driven feature engineering in structured datasets, where careful selection/construction can yield superior results compared to purely data-driven dimensionality reduction techniques. Finally, advanced learning-based models, including MLP+LSTM+Transformer and KAN+ResNet, deliver competitive results, particularly when handling more complex feature representations. However, their performance still falls short compared to simpler classifiers, which are easier to train and converge faster on this dataset. The primary reason lies in the nature of the dataset itself, which is binary-labeled with relatively well-defined, structured meteorological features. In such cases, simpler classifiers are inherently more effective as they can capture the underlying decision boundaries without the need for extensive training or hyperparameter tuning. On the contrary, advanced models often rely on complex architectures and large amounts of training data to realize their full potential, making them prone to overfitting or underfitting when applied to a smaller, structured dataset like this one. This demonstrates that, for problems with clear feature definitions and binary outputs, the additional complexity of advanced models may not translate to better performance and instead becomes an unnecessary computational burden.

\begin{table}[h!]
\centering
\resizebox{\textwidth}{!}{
\begin{tabular}{|l|ccccc|ccccc|ccccc|ccccc|}
\hline   
\multirow{2}{*}{\textbf{Method}} 
& \multicolumn{20}{c|}{\textbf{Feature Engineer Strategy}} \\
\cline{2-21}
& \multicolumn{5}{c|}{\textbf{\textit{Original Features}}} 
& \multicolumn{5}{c|}{\textbf{\textit{Selected + Constructed Features}}} 
& \multicolumn{5}{c|}{\textbf{\textit{Selected + Constructed + PC1 + PC2}}} 
& \multicolumn{5}{c|}{\textbf{\textit{Selected + Constructed + PC1 till PC8}}} \\
\cline{2-21}
 & \textit{Acc} $\uparrow$ & \textit{Prec} $\uparrow$ & \textit{Recall} $\uparrow$ & \textit{AUC} $\uparrow$ & \textit{F1} $\uparrow$ 
 & \textit{Acc} $\uparrow$ & \textit{Prec} $\uparrow$ & \textit{Recall} $\uparrow$ & \textit{AUC} $\uparrow$ & \textit{F1} $\uparrow$  
 & \textit{Acc} $\uparrow$ & \textit{Prec} $\uparrow$ & \textit{Recall} $\uparrow$ & \textit{AUC} $\uparrow$ & \textit{F1} $\uparrow$ 
 & \textit{Acc} $\uparrow$ & \textit{Prec} $\uparrow$ & \textit{Recall} $\uparrow$ & \textit{AUC} $\uparrow$ & \textit{F1} $\uparrow$ \\
 \hline
\multicolumn{21}{|c|}{\textbf{Non-learning Mathematical Approaches}} \\
\hline
LDA& 0.839& 0.698& 0.475& 0.848& 0.565& 0.733& 0.739& 0.721& 0.808& 0.730& 0.702& 0.706& 0.693& 0.761& 0.700& 0.711& 0.713& 0.706& 0.777& 0.709\\
Linear Regression& 0.837& 0.743& 0.400& 0.848& 0.520&  0.733& 0.739& 0.721& 0.808& 0.730& 0.702& 0.706& 0.693& 0.761& 0.700& 0.711& 0.713& 0.706& 0.777& 0.709\\
\hline
\multicolumn{21}{|c|}{\textbf{Weak Machine Learning Classifier}} \\
\hline
LASSO & 0.841& 0.713& 0.457& 0.845& 0.557&  0.735& 0.738& 0.728& 0.813& 0.733&  0.700& 0.701& 0.698& 0.759& 0.699 & 0.711& 0.712& 0.708& 0.775& 0.710 \\
ElasticNet & 0.841& 0.714& 0.457& 0.845& 0.557  & 0.732& 0.734& 0.728& 0.806& 0.731  & 0.700& 0.701& 0.698& 0.759& 0.699  & 0.711& 0.712& 0.708& 0.775& 0.710  \\
DT & 0.785& 0.508& \textbf{\textcolor[rgb]{0.004, 0.663, 0}{\underline{0.542}}}& 0.691& 0.524&   0.899&  \textbf{\textcolor{blue}{\uwave{0.850}}}& 0.970& 0.900& 0.906&   0.861&  \textbf{\textcolor{blue}{\uwave{0.826}}}& 0.914& 0.883& 0.868&   0.888&  \textbf{\textcolor{blue}{\uwave{0.836}}}& 0.966& 0.889& 0.896  \\
NB & 0.760& 0.453& 0.447& 0.698& 0.45&   0.703& 0.732& 0.641& 0.758& 0.684&   0.658& 0.753& 0.471& 0.726& 0.580&   0.683& 0.690& 0.666& 0.747& 0.678\\
KNN & 0.836& \textbf{\textcolor{red}{\underline{\underline{0.774}}}} & 0.357& 0.856& 0.488&   0.878& 0.812& \textbf{\textcolor{red}{\underline{\underline{0.982}}}}& \textbf{\textcolor{red}{\underline{\underline{0.989}}}}& 0.889&   0.845& 0.773& \textbf{\textcolor{red}{\underline{\underline{0.979}}}}& \textbf{\textcolor{red}{\underline{\underline{0.984}}}}& 0.864&   0.866& 0.798& \textbf{\textcolor{red}{\underline{\underline{0.981}}}}& \textbf{\textcolor{red}{\underline{\underline{0.987}}}}& 0.880\\
RF & \textbf{\textcolor{red}{\underline{\underline{0.852}}}} & \textbf{\textcolor{blue}{\uwave{0.750}}}& 0.488& \textbf{\textcolor{red}{\underline{\underline{0.875}}}}&  \textbf{\textcolor{blue}{\underline{\underline{0.591}}}}& \textbf{\textcolor{red}{\underline{\underline{0.938}}}}& \textbf{\textcolor{red}{\underline{\underline{0.910}}}}& 0.972& 0.938& \textbf{\textcolor{red}{\underline{\underline{0.940}}}}& \textbf{\textcolor{red}{\underline{\underline{0.896}}}}& \textbf{\textcolor{red}{\underline{\underline{0.849}}}}&  \textbf{\textcolor{blue}{\uwave{0.964}}}& \textbf{\textcolor[rgb]{0.004, 0.663, 0}{\underline{0.970}}}& \textbf{\textcolor{red}{\underline{\underline{0.903}}}}&  \textbf{\textcolor{red}{\underline{\underline{0.929}}}}& \textbf{\textcolor{red}{\underline{\underline{0.898}}}}& 0.968& \textbf{\textcolor[rgb]{0.004, 0.663, 0}{\underline{0.983}}}& \textbf{\textcolor{red}{\underline{\underline{0.932}}}}\\
LR & 0.841& 0.715& 0.458& 0.845& 0.559& 0.732& 0.734& 0.728& 0.806& 0.731& 0.700& 0.701& 0.698& 0.759& 0.699& 0.711& 0.712& 0.708& 0.775& 0.710\\
GB & 0.847& 0.690& \textbf{\textcolor{red}{\underline{0.548}}}&  \textbf{\textcolor{blue}{\uwave{0.866}}}& \textbf{\textcolor{red}{\underline{\underline{0.611}}}}&   0.817& 0.809& 0.831& 0.895& 0.820&   0.750& 0.740& 0.769& 0.825& 0.755&   0.801& 0.788& 0.824& 0.872& 0.806\\
DT+LR+RF & 0.834& 0.650& 0.508& 0.853& 0.570&   \textbf{\textcolor[rgb]{0.004, 0.663, 0}{\underline{0.914}}}& \textbf{\textcolor[rgb]{0.004, 0.663, 0}{\underline{0.871}}}&  \textbf{\textcolor{blue}{\uwave{0.973}}}&  \textbf{\textcolor{blue}{\uwave{0.973}}}& \textbf{\textcolor[rgb]{0.004, 0.663, 0}{\underline{0.919}}}&   \textbf{\textcolor[rgb]{0.004, 0.663, 0}{\underline{0.884}}}& \textbf{\textcolor[rgb]{0.004, 0.663, 0}{\underline{0.839}}}& 0.950& 0.946& \textbf{\textcolor[rgb]{0.004, 0.663, 0}{\underline{0.891}}}&   \textbf{\textcolor[rgb]{0.004, 0.663, 0}{\underline{0.904}}}& \textbf{\textcolor[rgb]{0.004, 0.663, 0}{\underline{0.857}}}&  \textbf{\textcolor{blue}{\uwave{0.969}}}& 0.967& \textbf{\textcolor[rgb]{0.004, 0.663, 0}{\underline{0.909}}}  \\
KNN+LR+RF & 0.843& \textbf{\textcolor[rgb]{0.004, 0.663, 0}{\underline{0.759}}}& 0.416& 0.863& 0.537&    \textbf{\textcolor{blue}{\uwave{0.903}}}& 0.849& \textbf{\textcolor[rgb]{0.004, 0.663, 0}{\underline{0.980}}}& \textbf{\textcolor[rgb]{0.004, 0.663, 0}{\underline{0.978}}}&  \textbf{\textcolor{blue}{\uwave{0.910}}}&    \textbf{\textcolor{blue}{\uwave{0.874}}}& 0.812& \textbf{\textcolor[rgb]{0.004, 0.663, 0}{\underline{0.973}}}&  \textbf{\textcolor{blue}{\uwave{0.964}}}&  \textbf{\textcolor{blue}{\uwave{0.885}}}&    \textbf{\textcolor{blue}{\uwave{0.890}}}& 0.832& \textbf{\textcolor[rgb]{0.004, 0.663, 0}{\underline{0.977}}}&  \textbf{\textcolor{blue}{\uwave{0.974}}}&  \textbf{\textcolor{blue}{\uwave{0.899}}}\\
\hline
\multicolumn{21}{|c|}{\textbf{Advanced Learning-based Model}} \\
\hline
MLP & 0.840& 0.685& 0.503& 0.849& 0.580  & 0.809& 0.783& 0.853& 0.876& 0.817  & 0.705& 0.712& 0.689& 0.765& 0.700  & 0.776& 0.757& 0.814& 0.849& 0.784  \\
CNN & 0.811& 0.679& 0.261& 0.789& 0.377  & 0.707& 0.730& 0.655& 0.775& 0.691  & 0.703& 0.720& 0.665& 0.764& 0.692  & 0.709& 0.730& 0.662& 0.779& 0.694  \\
GCN& 0.785& 0.558& 0.095& 0.690& 0.163  & 0.592& 0.609& 0.515& 0.627& 0.558  & 0.699& 0.708& 0.678& 0.759& 0.693  & 0.680& 0.707& 0.614& 0.745& 0.658  \\
ResNet & 0.840& 0.676& \textbf{\textcolor{blue}{\underline{0.520}}}& 0.848& 0.588  & 0.782& 0.798& 0.754& 0.865& 0.775  & 0.698& 0.730& 0.628& 0.762& 0.675  & 0.757& 0.762& 0.746& 0.835& 0.754  \\
RNN & 0.839& 0.737& 0.415& 0.843& 0.531  & 0.736& 0.737& 0.733& 0.810& 0.735  & 0.701& 0.707& 0.685& 0.760& 0.696  & 0.720& 0.728& 0.703& 0.790& 0.715  \\
LSTM & 0.840& 0.693& 0.483& 0.845& 0.570  & 0.789& 0.722& 0.820& 0.862& 0.795  & 0.759& 0.700& 0.706& 0.685& 0.695  & 0.802& 0.731& 0.738& 0.717& 0.727  \\
Transformer & 0.843& 0.703& 0.490& 0.855& 0.578  & 0.752& 0.765& 0.725& 0.829& 0.745  & 0.701& 0.706& 0.688& 0.761& 0.697 & 0.721& 0.733& 0.696& 0.792& 0.714  \\
KAN & 0.846& 0.735& 0.462& 0.856& 0.567& 0.748& 0.750& 0.745& 0.828& 0.747&   0.702& 0.708& 0.685& 0.763& 0.697&   0.730& 0.734& 0.721& 0.798& 0.727  \\
MLP+LSTM+Transformer & \textbf{\textcolor[rgb]{0.004, 0.663, 0}{\underline{0.850}}} & 0.720& 0.519& \textbf{\textcolor[rgb]{0.004, 0.663, 0}{\underline{0.869}}}& \textbf{\textcolor[rgb]{0.004, 0.663, 0}{\underline{0.603}}}& 0.802& 0.798& 0.809& 0.882& 0.804&   0.704& 0.710& 0.689& 0.764& 0.699&   0.783& 0.780& 0.788& 0.859& 0.784  \\
KAN+LSTM+Transformer & 0.847& 0.726& 0.483& 0.858& 0.580& 0.774& 0.776& 0.770& 0.858& 0.773 & 0.700& 0.708& 0.682& 0.762& 0.694& 0.747& 0.754& 0.734& 0.827& 0.744\\
KAN+ResNet & \textbf{\textcolor{blue}{\uwave{0.848}}} & 0.728& 0.485& 0.864& 0.582& 0.783& 0.796& 0.761& 0.866& 0.778& 0.704& 0.715& 0.677& 0.764& 0.695& 0.759& 0.764& 0.750& 0.839& 0.757\\
\hline
\end{tabular}
}
\caption{Quantitative experiment on model performances under different feature engineering strategies. \textit{\underline{Notes:}} Top 3 results are marked with \textcolor{red}{RED}, \textcolor[rgb]{0.004, 0.663, 0}{GREEN}, and \textcolor{blue}{BLUE}.}
\label{Quantitative experiment on model performances under different feature engineering strategies}
\end{table}

\section{Conclusion \& Future work} 
In this study, we explored the application of non-learning/learning-based and hybrid methodologies to enhance rainfall prediction performance, addressing challenges in meteorological variability and data complexity. Our systematic data pipeline within RAINER, including robust data preprocessing, feature engineering, and the integration of dimensionality reduction techniques such as PCA, provided high-quality inputs for predictive modeling. Comprehensive experiments with both weak and advanced classifiers revealed the potential of ensemble methods like voting to improve model performance, while parameter tuning via grid search further optimized accuracy. By evaluating models across diverse metrics, including precision, recall, accuracy, AUC, and F1 score, we gained insights into their strengths and limitations. These findings underscore the efficacy of machine learning approaches in capturing nonlinear dynamics and highlight the importance of feature construction for improved generalization. Our work contributes to bridging the gap between theoretical advancements and their practical applications in dynamic meteorological systems.

For future work, there are several promising directions to extend this research. First, we aim to apply the feature engineering pipeline developed in this study to other predictive domains, such as medical and healthcare datasets, to evaluate its effectiveness and generalizability across diverse applications. Second, we plan to tailor network architectures specifically for two-label prediction tasks or similar challenges, potentially proposing a novel architecture that aligns more closely with the unique requirements of these domains. Third, exploring interpretability techniques, such as SHAP values or attention mechanisms, may provide deeper insights into the decision-making process of machine learning models, aiding in model validation and adoption.

\bibliography{longforms,references}

\end{document}